%% file: main.tex
\documentclass[10pt,twocolumn,letterpaper]{article}

\usepackage[pagenumbers]{iccv} 
\usepackage{makecell}

\input{preamble}

\definecolor{iccvblue}{rgb}{0.21,0.49,0.74}
\usepackage[pagebackref,breaklinks,colorlinks,allcolors=iccvblue]{hyperref}

\def\paperID{21} 
\def\confName{ICCV}
\def\confYear{2025}

\title{Towards Large Scale Geostatistical Methane Monitoring \\ with Part-based Object Detection}

\author{  
Adhemar de Senneville\textsuperscript{1 *} \quad
Xavier Bou\textsuperscript{1 *} \quad
Thibaud Ehret\textsuperscript{2} \quad
Rafael Grompone\textsuperscript{1} \\
Jean-Louis Bonne\textsuperscript{3} \quad
Nicolas Dumelie\textsuperscript{3} \quad
Thomas Lauvaux\textsuperscript{3} \quad
Gabriele Facciolo\textsuperscript{1} \\
\textsuperscript{1}Université Paris-Saclay, CNRS, ENS Paris-Saclay, Centre Borelli, France \\
\textsuperscript{2}AMIAD, Pole Recherche, France \\
\textsuperscript{3}Université de Reims Champagne-Ardenne, France \\
\small\texttt{\href{https://adhemardesenneville.github.io/Large-Scale-Object-Detection/}{adhemardesenneville.github.io/Large-Scale-Object-Detection/}} 
}

\begin{document}
\twocolumn[{%
\renewcommand\twocolumn[1][]{#1}%
\maketitle%
\begin{center}
    \input{imgs/img_maps_grand_est}
\end{center}%
}]

\NoHyper
\setcounter{footnote}{1}
\footnotetext{* These authors contributed equally}
\endNoHyper


\input{sec/0_abstract}    
\input{sec/1_intro_sota}

\input{sec/2_methode}
\input{sec/3_experiments}

\input{sec/4_end}
{
    \small
    \bibliographystyle{ieeenat_fullname}
    \bibliography{main}
}

\input{sec/X_suppl}

\end{document}

%% file: preamble.tex
\usepackage{svg} 
\usepackage{graphicx} 
\usepackage{tikz}
\usepackage{caption}
\usepackage{pgfplots}
\usepackage{subcaption}
\usepackage{float}
\usepackage{pgffor}     
\pgfplotsset{compat=1.18}
\usepackage{comment}
\usepackage{tikz}
\usetikzlibrary{positioning}
\usepackage{rotating} 
\usepackage{multirow}

\newcommand{\AD}[2]{{\color{gray}#1}{\color{blue}#2}}


\renewcommand{\paragraph}[1]{\smallskip\noindent{\bf#1}}


%% file: imgs/img_maps_grand_est.tex
\begin{center}
\captionsetup{type=figure}
\centering

\begin{minipage}{0.6\linewidth} 
    \centering
    \begin{tikzpicture}
        \node[anchor=south west, inner sep=0] (image) at (0,0) 
            {\includegraphics[width=\linewidth]{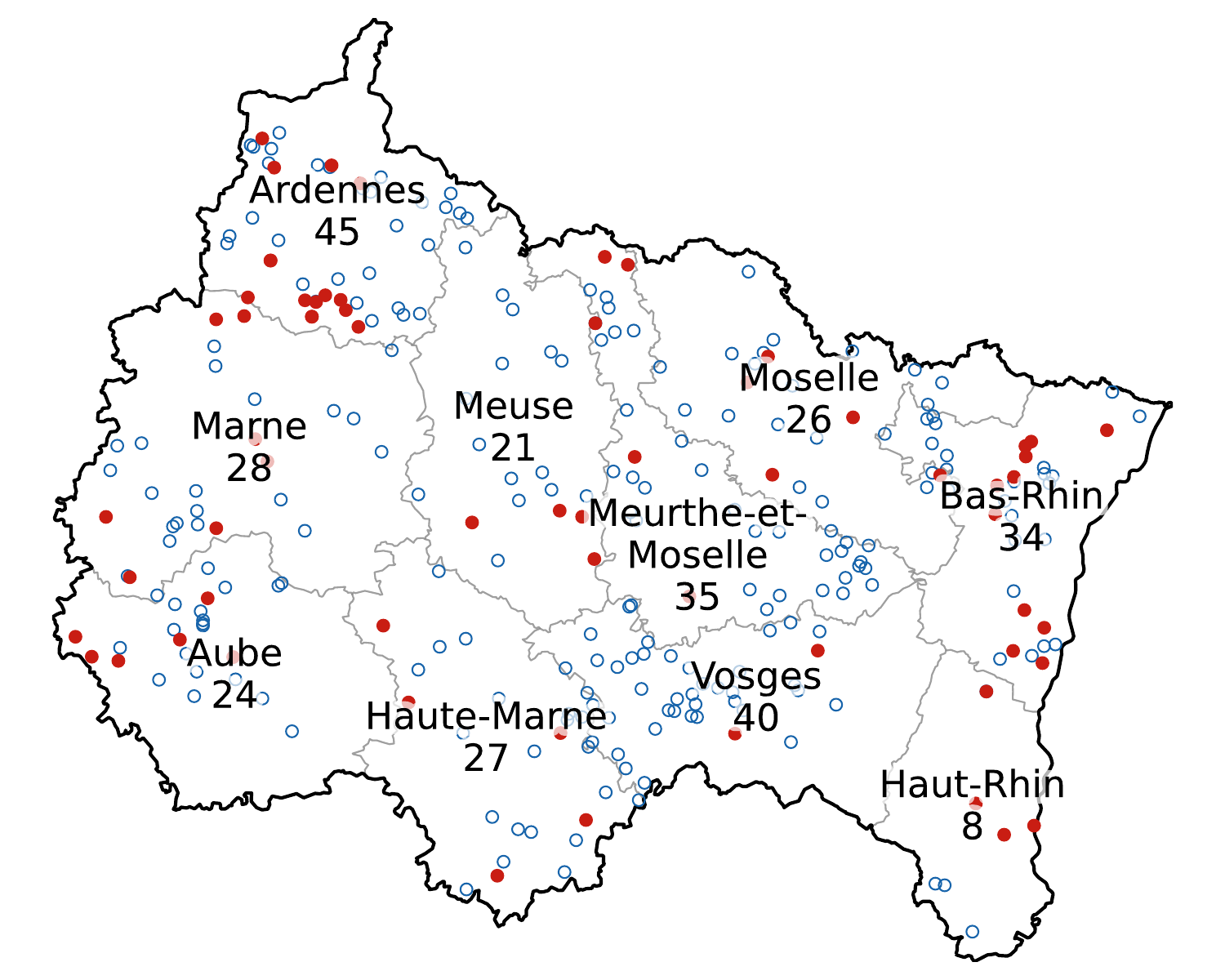}};
        \definecolor{myBlue}{rgb}{0.09, 0.393, 0.671} 
        \definecolor{myRed}{rgb}{0.787, 0.113, 0.073}
        \definecolor{myPurple}{rgb}{0.9, 0.4, 0.7}
        
        \small
        \node[fill=white, draw=none, rounded corners, inner sep=3pt] 
          at ([xshift=-110pt,yshift=-30pt]image.north east) {
          \begin{tabular}{l}
            \textcolor{myBlue}{\tikz{\node[draw=myBlue, fill=white, circle, minimum size=4pt, inner sep=0pt]{};}} Annotated Dataset \\
            \textcolor{myRed} {\tikz{\node[draw=none, fill=myRed,  circle, minimum size=4pt, inner sep=0pt]{};}} New detections
          \end{tabular}
        };           
        
    \end{tikzpicture}
\end{minipage}%
\hfill
\begin{minipage}{0.38\linewidth} 
    \centering
    \begin{tikzpicture}
        \definecolor{piles}{rgb}{1.0, 0.678, 0.118}    
        \definecolor{tanks}{rgb}{0.0, 0.271, 1.0}      
        \definecolor{biodigest}{rgb}{0.929, 0.090, 0.910} 
        
        \small
        \node[fill=white, draw=none, opacity=0, text opacity=1, rounded corners, inner sep=3pt] 
            at ([xshift=-50pt, yshift=50pt]image.north west) {
            \begin{tabular}{l}
                \textcolor{biodigest}{\textbf{\rule{10pt}{10pt}}} Bio-digester \,
                \textcolor{piles}{\textbf{\rule{10pt}{10pt}}} Piles \,
                \textcolor{tanks}{\textbf{\rule{10pt}{10pt}}} Tanks
            \end{tabular}
        };
    \end{tikzpicture}
    
    \begin{tabular}{cc} 
        \includegraphics[height=2.6cm, width=3cm]{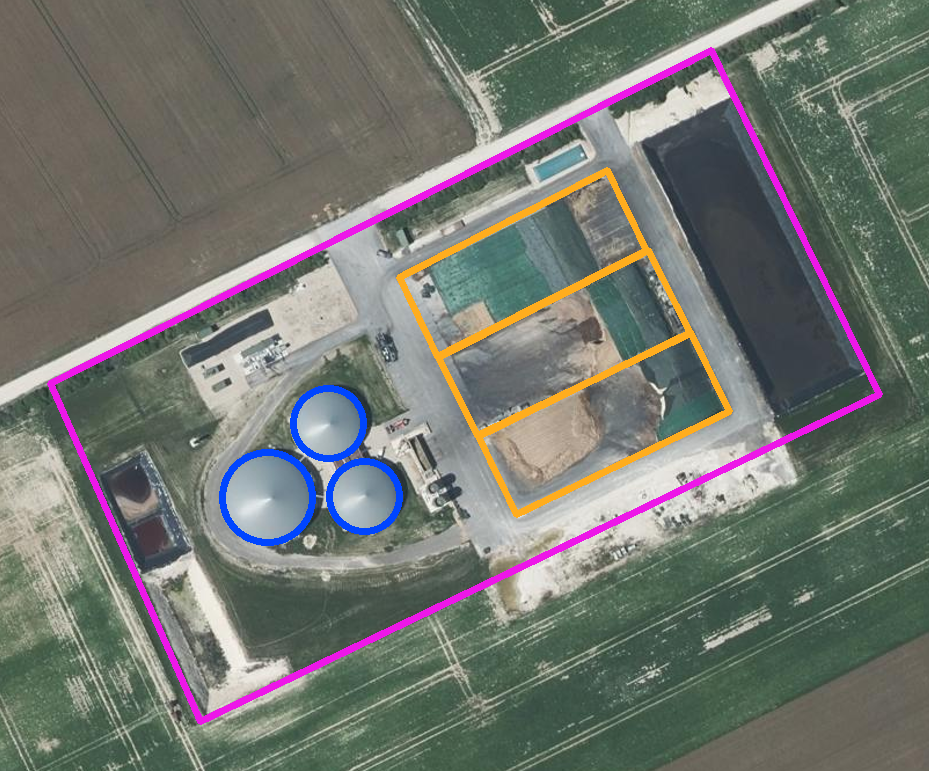} &
        \includegraphics[height=2.6cm, width=3cm, trim=0cm 1cm 0cm 1cm, clip]{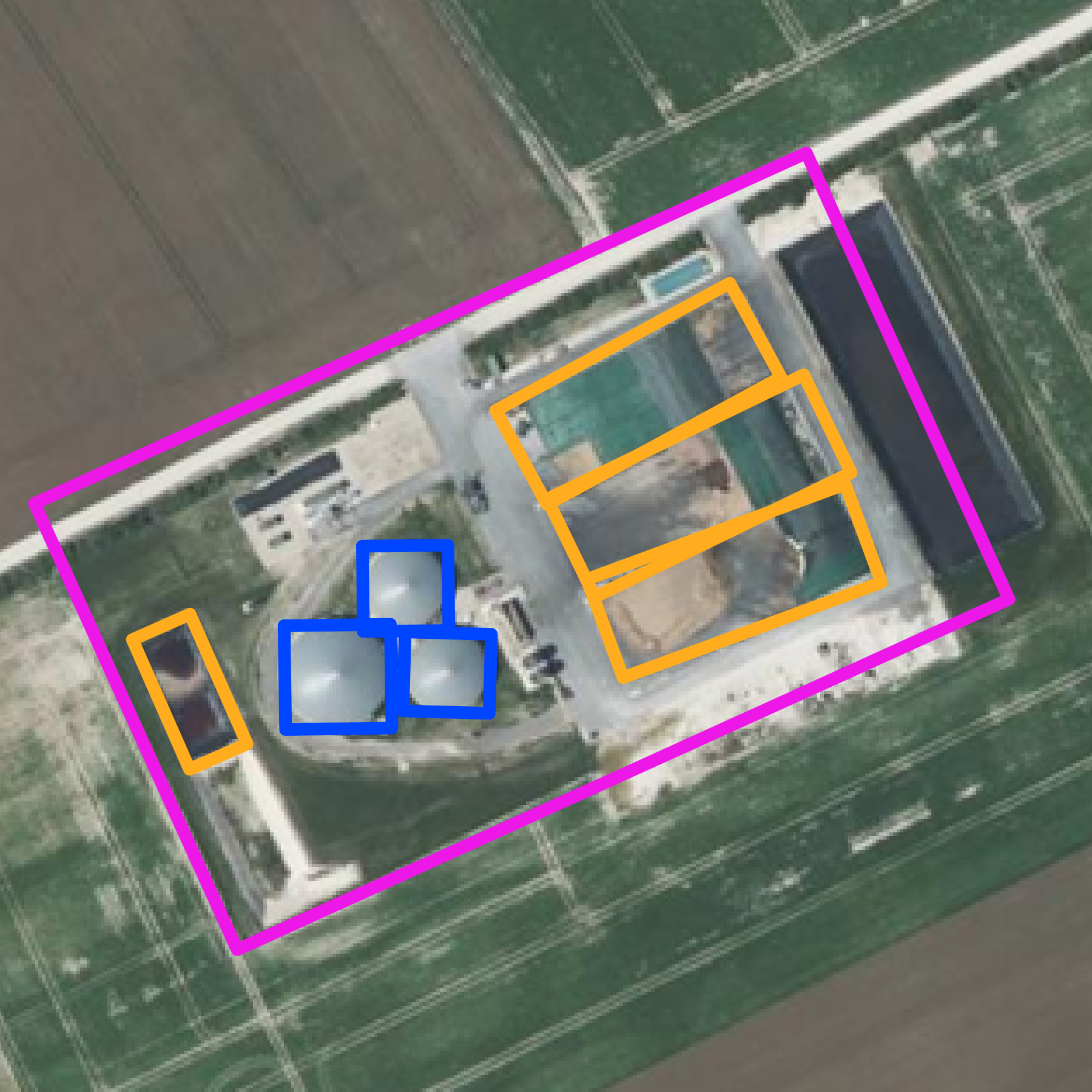} \\
        \includegraphics[height=3.2cm, width=3cm]{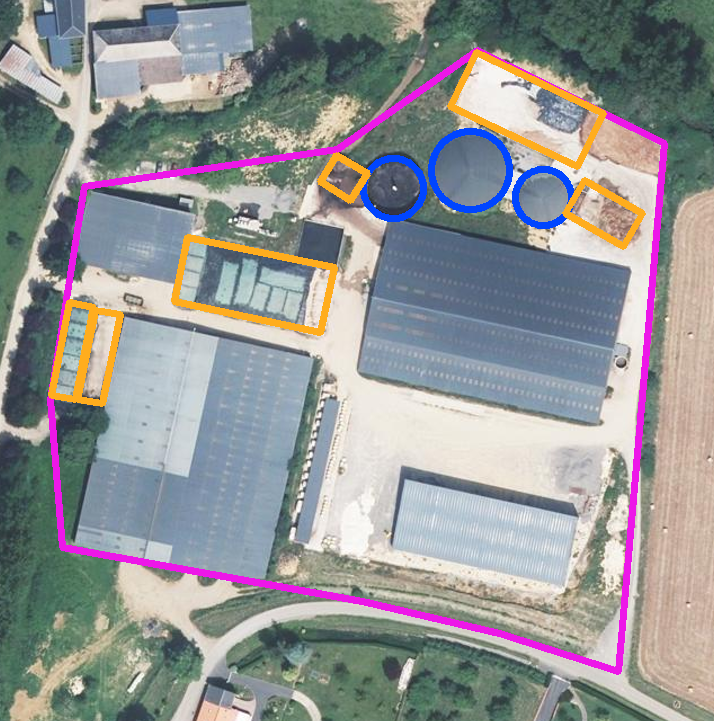} &
        \includegraphics[height=3.2cm, width=3cm]{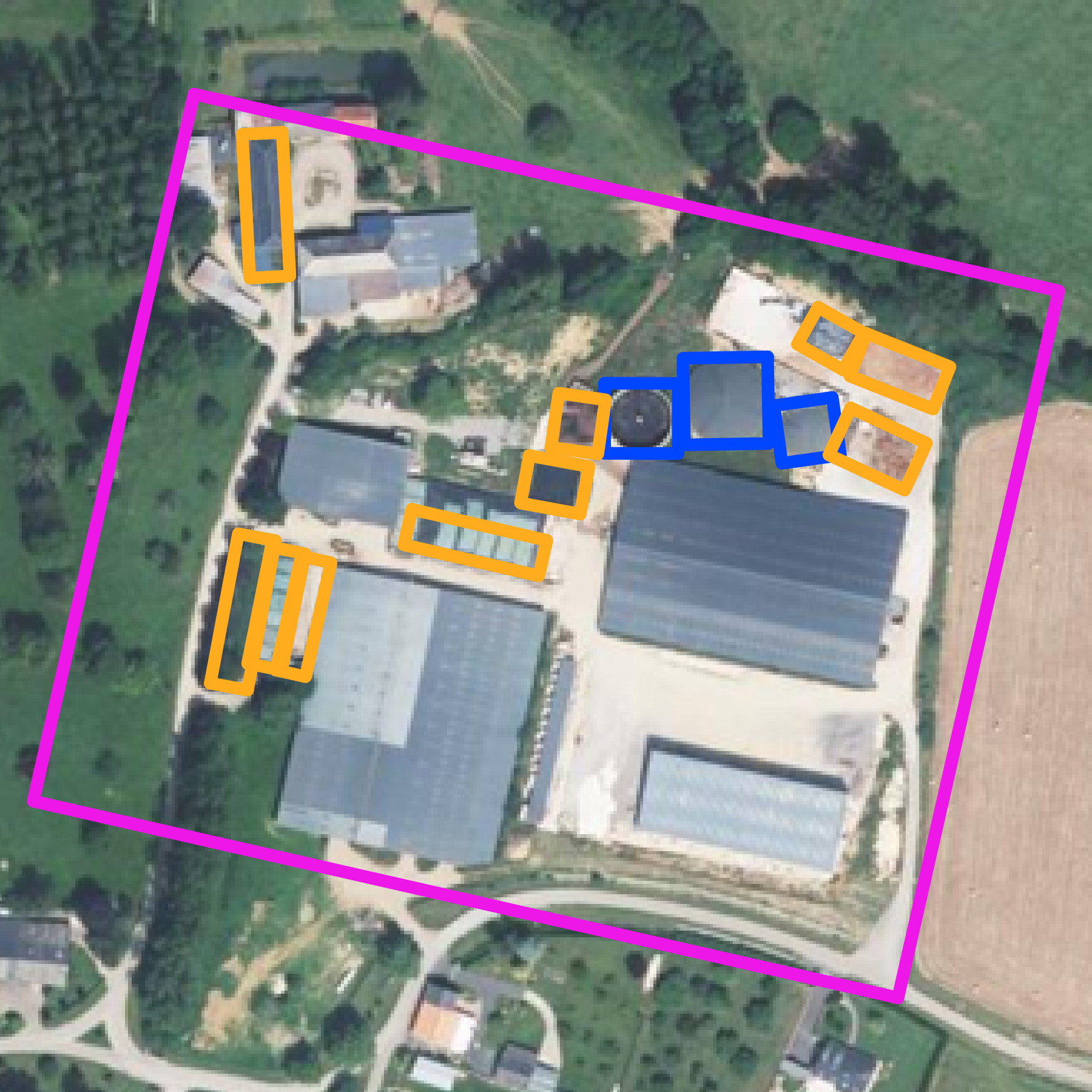} \\
         \textbf{(a)} & \textbf{(b)} \\
    \end{tabular}
\end{minipage}

\vspace{-6pt}

\caption{\textbf{Overview of our results in the French Grand Est region with the number of detected bio-digester sites in each department in 2023.} We use our model to detect unknown bio-digester sites in large areas. On the right, we show (a)~some examples of annotated bio-digester sites (from the validation set) with their sub-elements. (b)~Shows predictions from our model, even with detection errors and a small training set, the part-based detector reliably identifies bio-digester sites at scale.
}
\label{fig:maps_grand_est}
\end{center}

%% file: sec/0_abstract.tex
\begin{abstract}

Object detection is one of the main applications of computer vision in remote sensing imagery. Despite its increasing availability, the sheer volume of remote sensing data poses a challenge when detecting rare objects across large geographic areas. Paradoxically, this common challenge is crucial to many applications, such as estimating environmental impact of certain  human activities at scale. In this paper, we propose to address the problem by investigating the methane production and emissions of bio-digesters in France. We first introduce a novel dataset containing bio-digesters, with small training and validation sets, and a large test set with a high imbalance towards observations without objects since such sites are rare. We develop a part-based method that considers essential bio-digester sub-elements to boost initial detections. To this end, we apply our method to new, unseen regions to build an inventory of bio-digesters. We then compute geostatistical estimates of the quantity of methane produced that can be attributed to these infrastructures in a given area at a given time.
\end{abstract}


%% file: sec/1_intro_sota.tex
\section{Introduction}
\label{sec:intro}

Over an 80-year period, a methane ($CH_4$) molecule has a global warming potential approximately 80 times higher than that of carbon dioxide~\cite{Arias2021}.
As it has a shorter atmospheric lifespan and is estimated to be responsible for 30\% of the recent rise in global temperatures~\cite{IEA2022}, reducing $CH_4$ emissions is considered an effective strategy for mitigating near-term climate change~\cite{doi:10.1126/science.1210026}.

Due to the increasing availability of hyperspectral remote sensing data, recent efforts have focused on detecting and quantifying large methane emissions~\cite{pandey2019satellite,ouerghi2022automatic,elyes}. These approaches have proven effective in identifying significant methane leaks, enabling organizations to mitigate their impact on climate change. 
However, current large-scale remote sensing instruments are limited to large emitters~\cite{ehret2022global,pandey2019satellite,cusworth2022strong,thorpe2023attribution} in the order of tons per hour, lacking the sensitivity to detect small emitters. 
Tracking contributions from individual low emitters is challenging yet significant when aggregated into regional estimates. These emitters include biogas plants, wastewater treatment facilities, small community landfills, and oil and gas well pads.
Although a leak at a large methane-production facility will inevitably release a larger amount of $CH_4$, studies have shown that smaller sites tend to have a higher methane loss relative to their production~\cite{danish_report}.
Furthermore, a recent study reported that about 70\% of oil–gas methane emissions in the US originate from lower-emitting facilities~\cite{acp-25-1513-2025}.
Emissions from small sources are typically measured using airborne instruments. Nevertheless, aerial campaigns are expensive and limited in extent, thus country level aggregate emissions are derived through bottom-up approaches and self-reporting rather than direct measurement.
These bottom-up methane inventories are often non-publicly accessible or incomplete, introducing bias and under-reporting. 
Moreover, they lack the spatial and temporal resolution required to capture the evolution of emissions.

In contrast, some works in the remote sensing community leverage the broad coverage of satellite imagery to generate large-scale inventories of specific objects~\cite{PoultryInventory, GasInventory, Stowell2020}. This process, known as object inventorying, provides valuable insights into the spatial distribution of these objects over large areas and their potential environmental impact. 
Inspired by this paradigm, a bottom-up approach, based on localizing small methane-emitting sites, could serve as a feasible alternative for estimating large-scale emissions from low-emitters. 

Building on this idea, this article addresses agricultural biogas plants, which convert organic matter into methane through anaerobic digestion of organic waste, e.g. manure, crop residues, and food waste. As the number of biogas plants grows rapidly every year, existing inventories of bio-methane production quickly become incomplete and outdated. Consequently, the cumulative methane emissions from these facilities may represent a significant, yet poorly monitored, contribution to overall emissions. 

In this context, our work develops a bottom-up approach to detect small biogas plants in remote sensing images from a limited and incomplete database. We then use the resulting large-scale inventory to estimate aggregated methane production, from which a proportion of potential leaks can be inferred. Our method introduces a part-based probabilistic post-processing to control the precision of the detections at scale, enabling us to improve the model iteratively and identify a large number of bio-digesters across broader, unannotated regions of France. Moreover, we explore different sources of aerial and satellite imagery at different resolutions, and establish SPOT satellite data at 1.5\,m as an optimal trade-off between performance and efficiency. Qualitative and quantitative experiments demonstrate the generalization of our model to a different region in France. Our contributions can be summarized as:
\begin{itemize}
  \item We propose a methodology that, from a minimal dataset with sparse spatial distribution, yields a model that robustly detects bio-digesters nationwide.
  \item To the best of our knowledge, we release the first large-scale satellite dataset of bio-digesters, including facility and part annotations.
  \item We demonstrate that our methodology and dataset can be used to identify previously unknown bio-digester sites, update incomplete inventories or generate new ones. We evaluate the limits of our method in terms of spatial resolution, and provide visualizations of newly found, non-annotated sites. Lastly, we use bio-digester visual features to predict aggregated methane production of bio-digesters in Bretagne.
\end{itemize}

\section{Related works}
\paragraph{Object Detection.} 
Object detection localizes objects by predicting bounding boxes around them, typically defined by the center $(x, y)$ coordinates, the width, and the height~\cite{obj_det_review_1}. Object detectors also predict the class label and confidence score for every detection. Recent advances in deep learning have led to major improvements in object detection~\cite{obj_det_review_2}. Two-stage detectors like Faster R-CNN first propose object regions and then classify them~\cite{rcnn, fasterrcnn, maskrcnn}, typically offering higher accuracy at the cost of speed. In contrast, one-stage detectors perform localization and classification in a single step, enabling faster inference for large-scale imagery~\cite{cornernet, centernet, fcos, reppoints}, with the YOLO family being the most prominent~\cite{yolov1, yolov3, yolov7, yolov9, yolov11}. 
In remote sensing, oriented bounding boxes (OBBs) are commonly used to more accurately enclose objects~\cite{dota, fair1m, oriented_detection_1}. OBBs extend traditional bounding boxes by adding an angle parameter that specifies their rotation to better align with object orientation.
Consequently, many works have adapted object detection architectures to incorporate the orientation regression as well. Oriented R-CNN~\cite{Oriented_RCNN} extends the two-stage detectors of Faster RCNN by introducing an angle-sensitive region proposal network and specialized bounding box regression to accurately locate arbitrarily oriented targets. More recently LSKNet~\cite{li2024lsknetfoundationlightweightbackbone} employs a large selective kernel mechanism and enhances multi-scale feature extraction, leading to improved object detection performance in aerial imagery. Ding \textit{et al.}~\cite{rot_transf} adapted the transformer architecture for oriented object detection, and a wide range of other works propose to improve oriented object detection from the architectural point of view~\cite{Redet,R3det,fcos,gliding_vertex}, or mitigating boundary issues that come with the angular regression task~\cite{wasserstein_loss, kld, csl, psc, structure_tensor_obb, murrugarrallerena2025gauchogaussiandistributionscholesky}.

\paragraph{Object Detection in Remote Sensing.} The increasing availability of remote sensing images has led to a large amount of unlabeled data ready for its use~\cite{obj_det_survey_remote_sensing}. However, labeling data is costly and time-consuming. Fortunately, a number of remote sensing datasets have been proposed~\cite{LI2020296, dota, COWC_CarsOverheadWithContext, VEDAI}, driving significant research efforts toward the object detection problem~\cite{obj_rs_1,obj_rs_2,obj_rs_3, fsrw, bou2024exploringrobustfeaturesfewshot}. Well-known abundant datasets, such as DOTA~\cite{dota} or DIOR~\cite{LI2020296}, contain a wide range of categories e.g. \textit{car}, \textit{tennis court} or \textit{helicopter}. These datasets are often used to pre-train models before fine-tuning on smaller, remote sensing datasets, thus introducing aerial view information~\cite{huang2024mutdetmutuallyoptimizingpretraining, bou2023}.
Nevertheless, due to the nature of satellite images, these benchmarks contain notorious class imbalance, and we often see the same categories in all datasets~\cite{LI2020296}. To address this, other works have proposed datasets that focus on uncommon or fine-grained object categories. The SIMD~\cite{simd} dataset contains different types of planes and vehicles, such as \textit{propeller aircraft}, \textit{bus} or \textit{trainer aircraft}. Sun \textit{et al.}~\cite{SUN202150} introduce a Sewage Treatment Plant Dataset with one class: \textit{sewage treatment}. These have varying shapes, but more importantly they are composed of specific discriminative elements. The Ningbo Pylon dataset~\cite{Pylon_Datatset} introduce pylon observations in a sparsely distributed context, within a large amount of complex backgrounds. This closely resembles the realistic production environment of a model that searches instances in-the-wild.

\paragraph{Part-based object detection.}
Part-based object detection models \cite{felzenszwalb2009object, 4587597} represent objects as mixtures of multiscale deformable part models, where each part captures local appearance features, and their spatial relationships are modeled with constraints. This approach allows for robust detection of objects under variations in pose and occlusion. 
Although nowadays deep learning methods have surpassed traditional part-based models, the core ideas have been integrated in recent deep learning models.
DPM-CNN~\cite{DPM_CNN} integrates deformable part-based models with convolutional neural networks, replacing standard image features used in DPM with a learned CNN feature extractor. Part R-CNN~\cite{ZhangDGD14} extends region-based CNNs by incorporating explicit part-based reasoning, refining localization and enabling fine-grained recognition. PBNet~\cite{SUN202150} extends part-based detection concepts into deep learning for remote sensing, handling composite objects with diverse shapes and sizes. This approach has proved particularly relevant to detect objects that are comprised of smaller detectable parts.

\paragraph{Object Inventorying.} Several studies aim to build large-scale inventories of specific objects from aerial or satellite imagery for environmental monitoring.
For example, \cite{SolarPanelInventory} addresses the under-documentation of small-scale solar arrays by creating a dataset through large-scale crowdsourcing using OpenStreetMap. 
On the other hand, deep learning approaches have enabled efficient analyses over vast geographic areas.
Robinson \textit{et al.}~\cite{PoultryInventory} present a Deep Learning-based method to detect poultry barns from aerial imagery, producing an open-source national-scale poultry dataset. They demonstrated the model’s potential for large-scale environmental monitoring. 
Similarly, Ramachandran \textit{et al.}~\cite{GasInventory} propose a two-stage deep learning pipeline on high-resolution satellite imagery to detect oil and gas well pads and storage tanks. They identified over 70,000 unmapped well pads and more than 169,000 storage tanks, and demonstrated the potential of automated, nation-scale infrastructure mapping for improved methane emissions monitoring. 
More recently, Robinson \textit{et al.}~\cite{microsoft} built a global spatio-temporal dataset of commercial solar photovoltaic farms and onshore wind turbines using similar techniques and OpenStreetMap as labels.
To our knowledge, no attempt has been made to inventory small-scale methane production sites as bio-digesters. A work in this direction could aid large-scale monitoring of methane emissions, supporting sustainability efforts through spatial and temporal tracking of small emitters.

%% file: sec/2_methode.tex
\section{Data}
\label{sec:dataset}
This work strives for generalizable bio-digester detection from a limited set of annotations. To this end, we construct an initial small-scale dataset based on a few previously identified sites in France's Grand Est region.

\paragraph{Image source.} We extract images from DB ORTHO~\cite{bdortho}, a collection of RGB orthophotos at 20 cm per pixel resolution, acquired by the French national institute of geographic and forest information (IGN). We resample images to 50\,cm/pixel and generate crops of $1024 \times 1024$. In our spatial resolution experiments, we consider two additional image sources, namely SPOT6/7~\cite{Spot6-7_2023}, a commercial satellite RGB product at 1.5\,m per pixel, and Sentinel-2, a low-resolution platform providing images at 10\,m per pixel.

\paragraph{Annotations.} Ground truth labels are provided as geolocated segmentation annotations. For each bio-digester, we annotated three different classes: the whole bio-digester installation, the (anaerobic) digestion tanks, and biomass piles (feedstock storage areas). The latter two are smaller structures within bio-digester sites and are crucial for confirming site presence. Figure~\ref{fig:ign_spot} shows an example site with its annotations, for BD ORTHO (aerial), SPOT and Sentinel-2 data sources.
\input{imgs/ign_vs_spot_2}



\paragraph{Training and evaluation splits.}
We generate train and validation sets containing 163 and 40 examples, respectively. We also introduce 400 background images, i.e. images with no bio-digester, randomly selected and manually validated so that they contain buildings that are not bio-digesters. The dilution parameter $\alpha$, represents the fraction of non-background tiles in an image set. By diluting the annotated validation examples with background images, we can obtain a realistic idea of the number of false alarms produced by the detector, especially since we do not train with a large amount of examples. 
This step is essential to have a better control of the performance of our models and for the relevance of our ablation studies.

Moreover, we generate a test set to address the capabilities of finding new exemplars in large-scale areas. To this end, we gather 5096 images covering the entire department of Marne, which is one of the 101 departments of France. Initially, 21 bio-digesters were present in the department of Marne to our knowledge. Figure~\ref{fig:grand_est_map} illustrates the geospatial distribution of train and validation sets, and
Table~\ref{tab:dataset_summary} provides a summary of the dataset splits used for the study.

\input{maps/dataset_map}
\begin{table}
  \centering
  \setlength{\tabcolsep}{4pt}
  \begin{tabular}{@{}lccccc@{}}
    \toprule
    Dataset  & Number of & Annotated vs & $\alpha$  \\
            & Images   & Background Tiles &   \\
    \midrule
    Training   & 326    & 163 vs 163   & 50\%    \\
    Validation & 440    & 40 vs 400    & 9.1\%     \\
    \midrule
    Test - Marne& 5096  & 27 vs 5069   & 0.53\%      \\
    \bottomrule
  \end{tabular}
  \caption{\textbf{Summary of the proposed bio-digester dataset} used in the study and the ratio of annotated tiles $\alpha$ in the different sets. Note that in test environment, 27 tiles contain the 21 bio-digesters (due to tile overlap some sites appear in multiple tiles).
  }
  \label{tab:dataset_summary}
\end{table}

\begin{figure*}[t]
    \centering
    \includegraphics[width=1\linewidth,trim={0.5cm 5.85cm 0cm 0cm},clip]{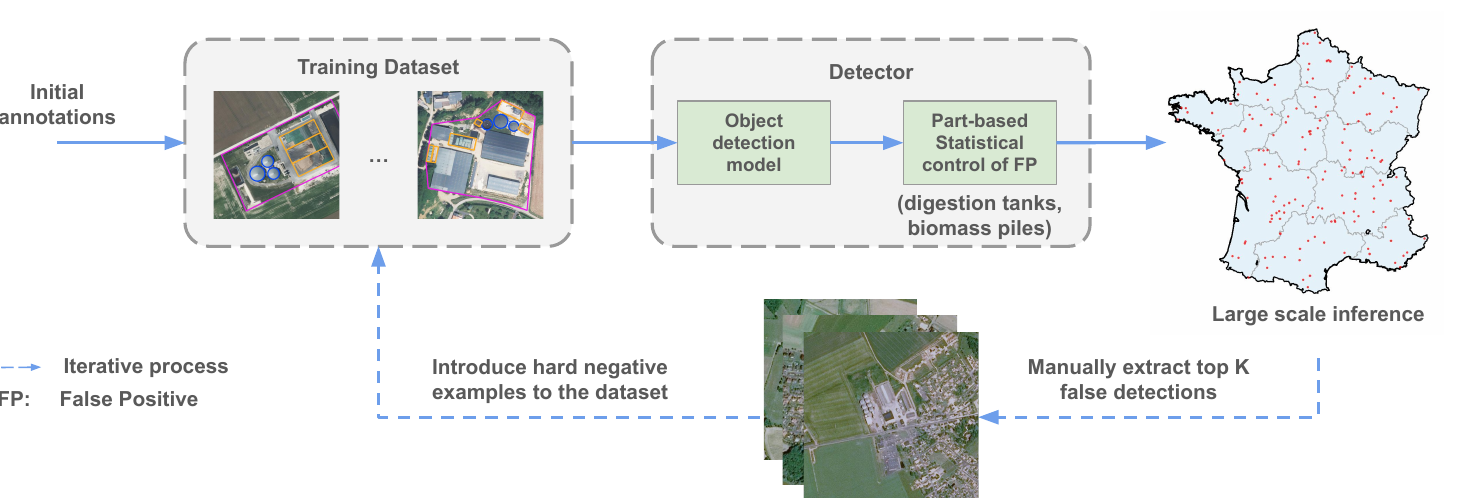}
    
    
    \caption{\textbf{General diagram of the proposed approach}. The initially annotated dataset is used to train a conventional object detector network. Then, we apply a part-based statistical method to boost detection performance, making large-scale detection bearable. Subsequently, an iterative process takes place, where 1) the trained detector is applied over larger, diverse regions and 2) the top $K$ detections are manually verified and introduced to the dataset. The same process is then conducted with the resulting larger dataset.
    }
    \label{fig:detector_inference}
\end{figure*}

\paragraph{Large-scale dataset.} In this work, we iteratively detect and manually validate bio-digester sites in previously unlabeled areas, identifying newly built sites that are absent from existing databases. Moreover, as part of  the iterative refinement, we introduce hard negative examples, i.e. examples that tend to be false positive detections, so that the network learns to ignore them. As a result, we construct—to the best of our knowledge—the first large-scale bio-digester detection dataset. We strongly believe it can pave the way to advance the field and support future research on methane estimation from low-emitting sources. The resulting dataset includes new detections that were not labeled in Grand Est and Bretagne regions as well as 200 hard negative samples. More information on the final database and its process is provided in Section~\ref{sec:experiments}.

\section{Method}
\label{sec:method}

We aim to train a model from a limited set of annotations that is able to generalize across different regions. To this end, our methodology first trains a conventional object detector on the set of available annotations. Then, we leverage the annotated parts to discriminate true from false positives, and reducing the number of false detections. Lastly, we iteratively apply the detector on large non-annotated areas and manually 
introduce the most confident false positives to the dataset to improve robustness, and repeat the training procedure iteratively. Figure~\ref{fig:detector_inference} illustrates this process, and each step is detailed in the upcoming paragraphs.

\paragraph{Baseline detection.} We first train an object detector on the initially annotated dataset described in Section~\ref{sec:dataset}. To do so, we pre-train the network on DOTA~\cite{dota}, which contains a large amount of satellite images. Then, we fine-tune the model to predict the three different categories annotated on the bio-digester dataset—the site itself, the digestion tanks, and the biomass piles. We consider several architectures and compare them later in Section~\ref{sec:experiments}, i.e. YOLOv8~\cite{yolov8}, Faster R-CNN~\cite{fasterrcnn}, Oriented R-CNN~\cite{Oriented_RCNN} and LSKNet~\cite{li2023largeselectivekernelnetwork}. We set this simple supervised model as the detection baseline, where after inference over large areas, we extract all bio-digester detections that meet a predefined confidence threshold.
Naturally, in such low data regime, a low threshold gives high recall and low precision, while a high threshold gives good precision and low recall. Although the detector has relatively good performance on the validation set, in large areas, the number of false alarms will increase due to the number of background images and the presence of unseen objects. Hence, we leverage essential parts of bio-digester sites and propose to remove false alarms taking inspiration from part-based models~\cite{felzenszwalb2009object}.

\paragraph{Part-based identification of false alarms.}  
In this section we use part detectors to boost the performance of the final detector. Although each detector independently produce high amounts of false positives, combining such weak detectors alike in boosting~\cite{FREUND1997119, schapire2003boosting} into a unified strong classifier allow to improve detection performance.

Humans naturally identify fine-grained objects like bio-digesters by first spotting potential candidates, then confirming them by checking for characteristic parts—such as tanks and piles, as shown in Figure~\ref{fig:maps_grand_est}.
Following this intuition, we define the detection confidence based on the presence of these parts within the detected bio-digester.
Specifically, 
{using our detector we count the number of biomass piles and digestion tanks inside the facility bounding box.} 

Let \( \mathbf{p} = (p_1, p_2, \dots, p_{N_{max}}) \) be the vector of probabilities of all detections of an object class where $N_{max}$ is the total number of extracted detections of that class inside a given bio-digester site. Let \( \mathbf{p_t} \) and \( \mathbf{p_p} \) be respectively the vectors of detection probabilities for tanks and piles, and  \( N_t \) and \( N_p \) denote the number of detected tanks and piles, respectively. From that we can compute the probability of having $N_{object}$ --where object is either tank or pile-- inside the bio-digester site as 
\begin{equation}\label{eq:n_obj}
p(N_{object} \!=\! N | \mathbf{p_\text{object}} ) = \!\!\!\!\!\!\!\sum_{\substack{I \subseteq \{1, \dots, N_{max}\}\\ |I| = N}} \, \prod_{i \in I} p_i \prod_{j \notin I} (1 - p_j).
\end{equation}
In Equation~\eqref{eq:n_obj}, we assume that each detection is modeled as an independent Bernoulli trial with success probability \( p_i \). Therefore, the probability of observing exactly \( N \) true detections given the vector of detection probabilities \( \mathbf{p}_{\text{object}} \) is computed by summing over all combinations of successes (with probability \( p_i \)) and failures (with probability \( 1-p_i \)).

We then estimate the prior  distribution of the number of tanks and piles within a bio-digester. Figure~\ref{fig:comparison} shows histograms of tanks and piles per site in the training data. All bio-digesters in the dataset contain at least one tank, with three tanks being the most common. We approximate the probability $p(D \mid N_t, N_p)$ that a site is a bio-digester given the detected number of tanks $N_t$ and piles $N_p$ inside it. To do so, we consider two approaches: (a)~modeling with a bivariate Poisson distribution, and (b)~using the empirical histogram under the assumption of independence (as the two variables exhibit low correlation).

Given a detection and a number of tanks and piles we can now estimate the probability of the detection being a bio-digester, where \( p_{b} \) is the bio-digester detection confidence. Given the Law of total probability, the detection score $p(D| \mathbf{p_t},\mathbf{p_p},p_{b})$ is then computed as
\begin{align}
p(D|\mathbf{p_t},\mathbf{p_p},p_{b}) = p_b \sum_{N_t,\,N_p \geq 0}& p(N_t \mid \mathbf{p_t})\, \cdot p(N_p \mid \mathbf{p_p}) \nonumber \\
&\quad \!\!\!\!\!\!\!\! \cdot \, p(D \mid N_t, N_p),
\end{align}
which corresponds to the overall probability of a detection given the observed bio-digester, tanks and piles scores.

Note that one could count the number of tanks and piles after applying a detection threshold; however, this would add a difficult-to-tune parameter to our detection pipeline. 

\input{data/histograms}

\paragraph{Iterative improvement.} 
Bio-digester sites show visual similarity to other structures, such as industrial zones, oil storage areas, or farms. This favors high false positive rates and compromises the reliability of detections. Despite the limited annotation starting point, a large amount of unlabeled territory can be leveraged to acquire new samples. Hence, we use the pre-trained model and the part-based statistical false alarm detector to improve the detector based on false positive detections. Taking inspiration from~\cite{microsoft}, we identify high-confidence false positive detections over large land areas via human inspection of the top $K$ most confident detections. These hard examples are added back into the training set without any annotation as new backgrounds at each iteration. Similarly, a new \textit{harder} validation set is built to evaluate against these challenging examples at each iteration as well. Introducing these confident false positives, also known as hard negative examples, should yield the detector more robust to challenging observations and provide a more realistic evaluation of the performance of the method.

Moreover, we considered adding correct \textit{in-the-wild} detections to increase the number of training sites. However, early experiments showed that, since most bio-digester sites are visually similar, adding new positive examples that already fit the learned distribution provided limited benefit. Instead, teaching the network what is not a bio-digester proved to be a more impactful and cost-effective alternative.

We conduct a total of 3 iterations including the first initial training. Each time we inspect all unlabeled detections in the test set and introduce the $K=100$ most confident false positives to the training set.

%% file: imgs/ign_vs_spot_2.tex
\newcommand{\igncrop}{0pt}  
\newcommand{\spotcrop}{0pt}   

\begin{figure}
    \centering
    \begin{tabular}{@{}cc}
        \subcaptionbox{Aerial (0.5m)}{
            \includegraphics[trim=\igncrop{} \igncrop{} \igncrop{} \igncrop{},clip,width=0.45\linewidth]{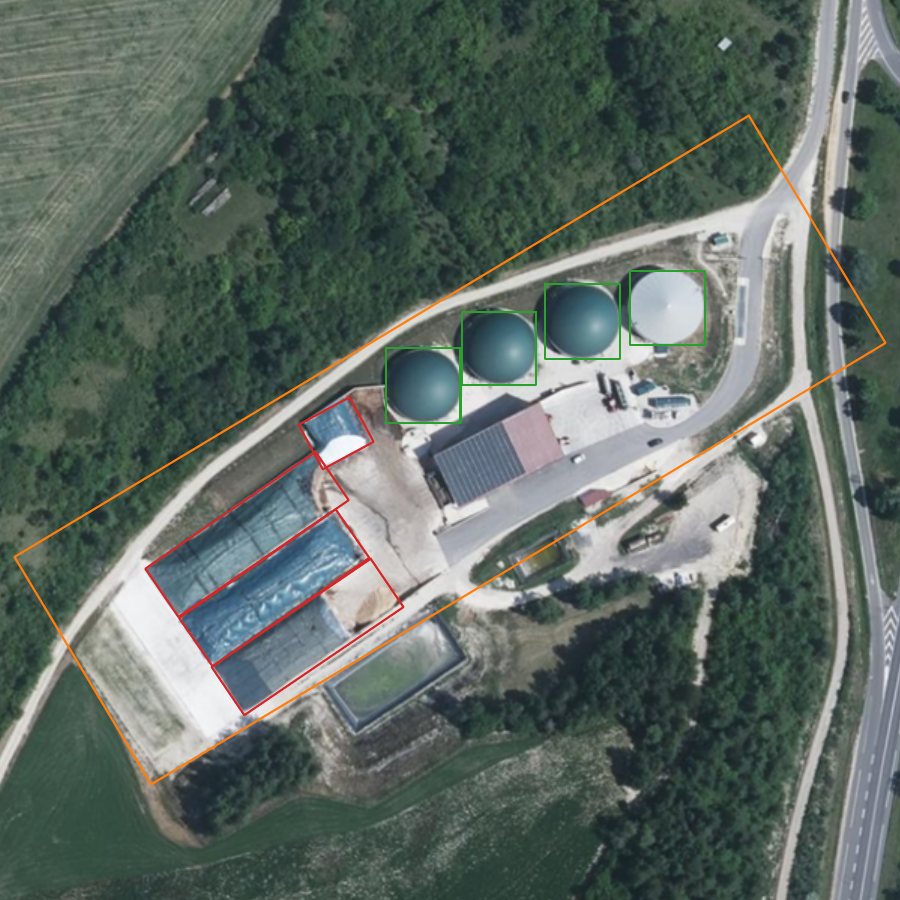}
        } &
        \subcaptionbox{Aerial (1.5m)}{
            \includegraphics[trim=\spotcrop{} \spotcrop{} \spotcrop{} \spotcrop{},clip,width=0.45\linewidth]{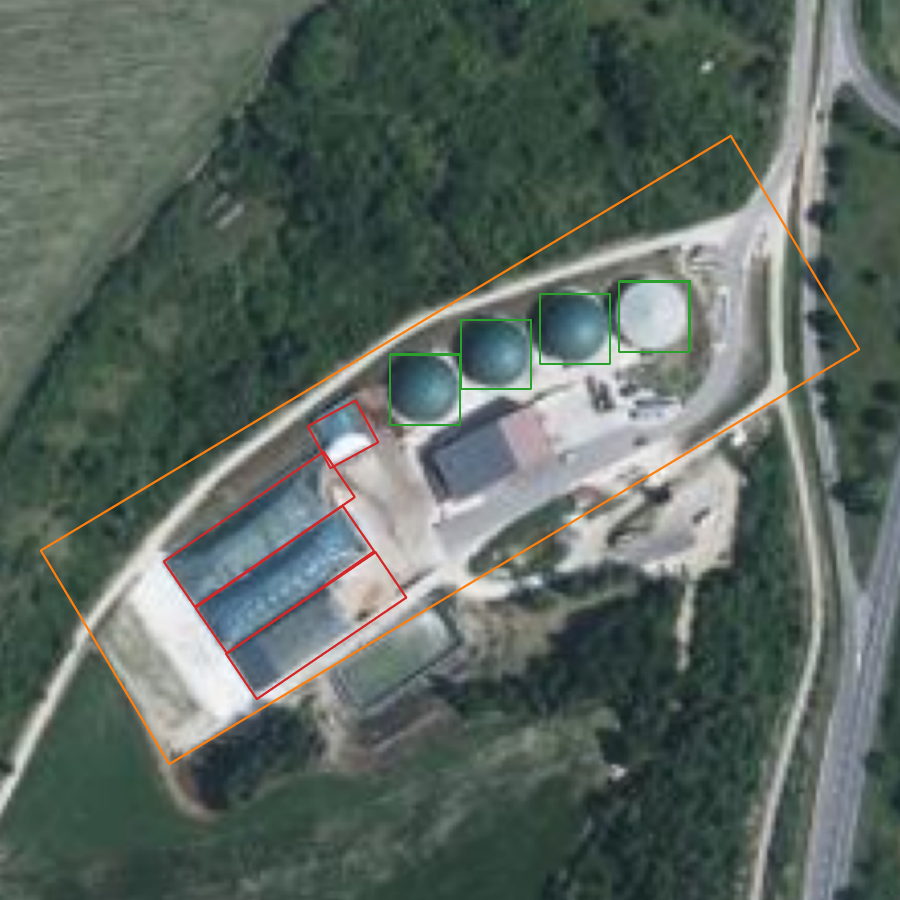}
        } \\
        \subcaptionbox{Sentinel (10m)}{
            \includegraphics[trim=\igncrop{} \igncrop{} \igncrop{} \igncrop{},clip,width=0.45\linewidth]{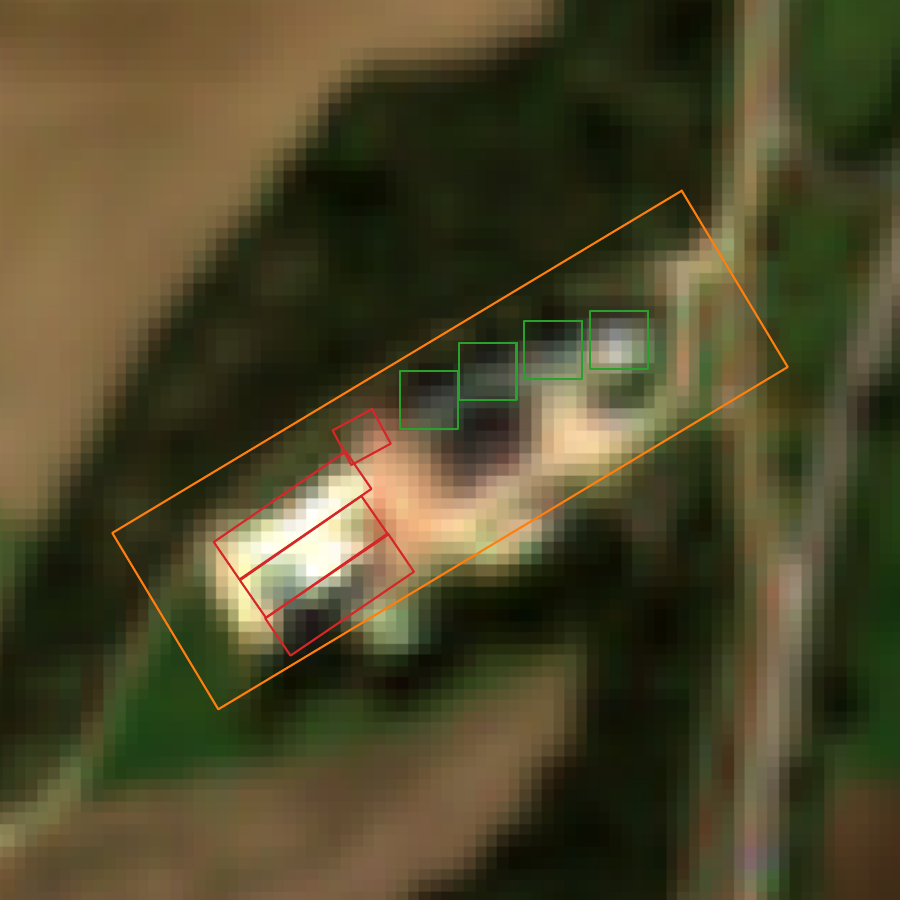}
        } &
        \subcaptionbox{SPOT (1.5m)}{
            \includegraphics[trim=\spotcrop{} \spotcrop{} \spotcrop{} \spotcrop{},clip,width=0.45\linewidth]{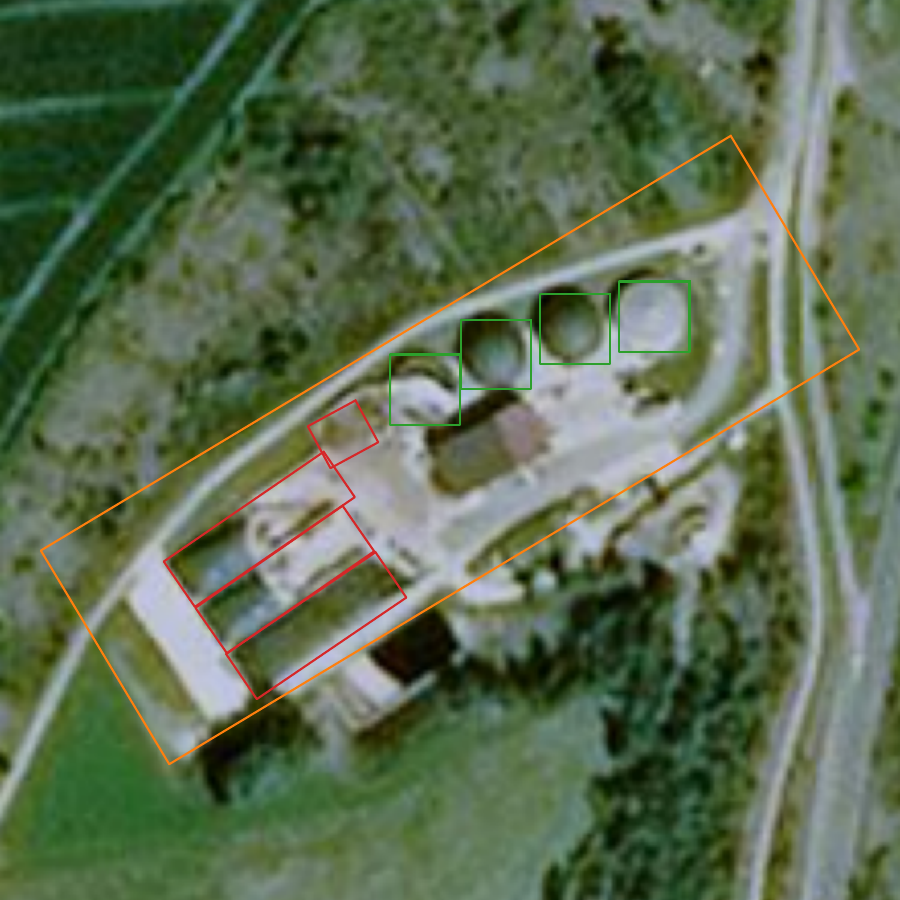}
        } \\
    \end{tabular}

\vspace{-3pt}
    
    \caption{\textbf{Illustration of aerial and satellite datasets.} with their annotations. (a) and (b) correspond to BD ORTHO images resampled at 0.5 and 1.5m per pixel, respectively. Sentinel-2 (c) and SPOT (d) images are shown on the bottom.
    }
    \label{fig:ign_spot}
\end{figure}

%% file: maps/dataset_map.tex
% ...

\begin{figure}[t]
    \centering
    \begin{tikzpicture}
        \node[anchor=south west, inner sep=0] (map) at (0,0) {\includegraphics[width=0.3\textwidth]{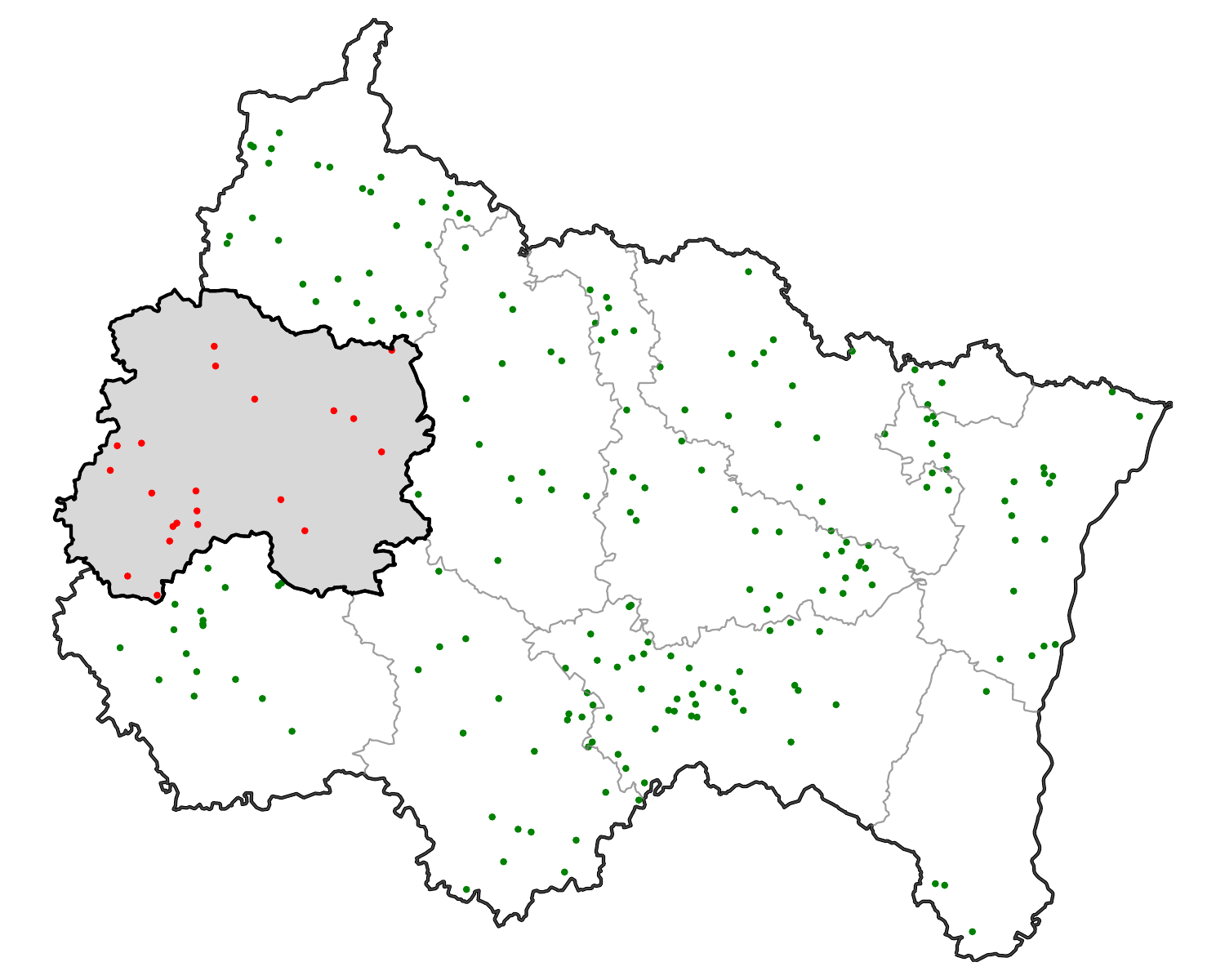}};

        \small
        \node[anchor=north east, xshift=50pt] at (map.north east) {
            \begin{tabular}{ll} 
                    \textcolor[rgb]{1.00,0.00,0.00}{\textbullet} & \hspace{-2pt} Test Set \\
                    \textcolor[rgb]{0.00,0.50,0.00}{\textbullet} & \hspace{-2pt} Train–Val Set \\
            \end{tabular}
        };
    \end{tikzpicture}

\vspace{-3pt}

    \caption{\textbf{Map of the Grand Est region of France.} It includes the test set locations in red only present in the department of Marne. The train-val set locations in green are located in the rest of the Grand Est region.}
    \label{fig:grand_est_map}
\end{figure}

%% file: data/histograms.tex
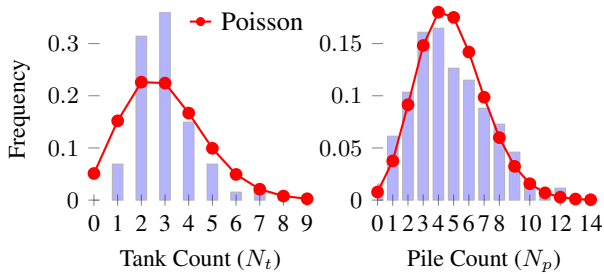
\begin{figure}[!t]
    \raggedleft
    %
    \begin{subfigure}[b]{0.4\columnwidth}
        \centering
        \begin{tikzpicture}[baseline, trim left]
            \begin{axis}[
                scale only axis,
                axis line style={draw=none},
                axis x line=bottom, 
                axis y line=left,
                width=\linewidth*0.85,
                height=\linewidth*0.75,
                symbolic x coords={0,1,2,3,4,5,6,7,8,9}, 
                xtick=data,
                ymin=0,
                ylabel={Frequency},
                ylabel style={at={(-0.35,0.8)}, anchor=east},
                xlabel={Tank Count ($N_t$)},
                legend style={font=\small, at={(0.98,0.98)}, anchor=north east},
                bar width=4pt,
                enlarge x limits=-5,
                tick label style={font=\small},
                label style={font=\small},
            ]
                \addplot[ybar, fill=blue, opacity=0.3] 
                    coordinates {(0,0) (1,0.06897) (2,0.31418) (3,0.36015) (4,0.14943) (5,0.06897) (6,0.01533) (7,0.01916) (8,0.00383) (9,0.00)};
                
                \addplot[mark=*, mark size=2pt, red, thick, solid] 
                    coordinates {(0,0.05106) (1,0.15194) (2,0.22608) (3,0.22425) (4,0.16683) (5,0.09929) (6,0.04925) (7,0.02094) (8,0.00779) (9,0.00257)};

                \draw[red, thick] (axis description cs:0.45,0.95) 
                      -- (axis description cs:0.55,0.95)
                      node[anchor=west, black] {Poisson};
                \filldraw[red] (axis description cs:0.50,0.95) circle (1.5pt);
            \end{axis}
        \end{tikzpicture}
        \label{fig:tanks}
    \end{subfigure}
    \hspace{3mm}
    \begin{subfigure}[b]{0.4\columnwidth}
        \centering
        \begin{tikzpicture}[baseline, trim left]
            \begin{axis}[
                scale only axis,
                axis x line=bottom, 
                axis y line=left,
                axis line style={draw=none},
                width=\linewidth*0.85,
                height=\linewidth*0.75,
                symbolic x coords={0,1,2,3,4,5,6,7,8,9,10,11,12,13,14}, 
                xtick={0,1,2,3,4,5,6,7,8,10,12,14},
                ymin=0,
                xlabel={Pile Count ($N_p$)},
                legend style={font=\small, at={(0.98,0.98)}, anchor=north east},
                bar width=4pt,
                enlarge x limits=0.0,
                tick label style={font=\small},
                label style={font=\small},
                yticklabel style={/pgf/number format/fixed}, 
            ]
                \addplot[ybar, fill=blue, opacity=0.3] 
                    coordinates {(0,0.01533) (1,0.06130) (2,0.10345) (3,0.16092) (4,0.16475) 
                 (5,0.12644) (6,0.11494) (7,0.08812) (8,0.07280) (9,0.04598) 
                 (10,0.01916) (11,0.00766) (12,0.01149) (13,0.00383) (14,0.00383)};
                \addplot[mark=*, mark size=2pt, red, thick, solid] 
                    coordinates {(0,0.00774) (1,0.03761) (2,0.09143) (3,0.14819) (4,0.18012) 
                 (5,0.17516) (6,0.14194) (7,0.09859) (8,0.05992) (9,0.03237) 
                 (10,0.01574) (11,0.00696) (12,0.00282) (13,0.00105) (14,0.00037)};
            \end{axis}
        \end{tikzpicture}
        \label{fig:piles}
    \end{subfigure}

    \caption{\textbf{Histogram of per-site number of digestion tanks and biomass piles in the train set.\vspace{-10pt}}}
    \label{fig:comparison}
\end{figure}

%% file: sec/3_experiments.tex
\section{Experiments}
\label{sec:experiments}

\paragraph{Implementation details.} Due to the elongated and closely packed nature of biomass piles, we consider oriented object detectors in addition of traditional object detectors. This way, non-maximum suppression becomes less problematic for overlapping diagonally arranged detections. We train our models over 100 epochs, using a batch size of 2 and Adam optimizer ($\beta_1$ = 0.9, $\beta_2$ = 0.999, $\epsilon$ = 1\text{e}-8) with a weight regularization of 0.0001. We apply data augmentation consisting of horizontal, vertical, and diagonal flipping. For each model, we use the learning rate scheduler and loss configurations from the original implementations. Based on our experiments, we selected LSKNet as our final object detector. All experiments were conducted on a single NVIDIA  P5000 GPU.

\paragraph{Metrics.} Bio-digester sites often lack precise spatial boundaries, making traditional IoU thresholds unsuitable for evaluation. Standard IoU-based metrics frequently misclassify correct detections, particularly when peripheral structures such as buildings are variably included or excluded from the detection bounding box.
We instead use a distance-based rule: if the center of a detection lies within 200\,m of any annotated site, it is considered a true positive. This threshold corresponds to the maximum offset observed among true positives in our data. If a site is detected multiple times even after non-maximum suppression, we consider that as one detection.
$AP_{dist}$ then refers to average precision computed using the distance-based matching criterion instead of the standard IoU threshold. 
To evaluate the detector for a given class we report the standard IoU‐based metrics $AP_{50}$, the average precision at an IoU threshold of 0.5. We chose a threshold of 0.5 as exactly positioning a given object is not essential to our analysis compared to just detecting it. Additionally $mAP_{50}$ is the mean average precision across all classes.


\paragraph{Results on the Grand Est Region.}
We extract the BD ORTHO aerial data for the entire Grand Est region, resample it at 1.5\,m, and  run inference with our model --processing over 36000 images in about 5 hours. Average-precision curves are shown in Figure~\ref{fig:precision_grand_est}. 
As seen, the part-based statistical approach significantly improved the performance.
We observe a discrepancy in the precision at low recall. Upon inspection of false alarms, we found that about 63 bio-digester sites were not referenced in our database, which represents 25\% of our detections. 
The geospatial distribution of the newly found bio-digesters are shown in red in Figure~\ref{fig:maps_grand_est}, where blue represents previously known positions.

\input{imgs/ap_grand_est}

\paragraph{Generalization to other locations.}
To evaluate the generalization capabilities of our method, we select two regions of France that were not used during training: Marne and Bretagne.
The Marne region includes 21 known bio-digesters, of which 18 were correctly identified, and 3 were missed. Additionally, 10 new sites were detected that were not present in the reference database, demonstrating the model's ability to generalize beyond known data. Moreover, Bretagne is known for its abundant bio-digestion sites and a database is publicly available~\cite{AILE2025}. We observe a decrease in performance in the recall; however, results are still relevant with a precision of 70\% and 35 out-of-database correct detections in the region.
These results are summarized in Table~\ref{tab:detection_results}, showing the generalization from the Grand Est training region to the two other regions in France.

\input{data/normandie_bars_2}

\paragraph{Part-based detection.} 
We conduct an ablation of the proposed baseline and part-based approaches to minimize the number of false positive detections. To this end, we provide in Table~\ref{tab:fp_ablation} the results achieved for each case in the validation set. Although the Poisson distribution takes into account the correlation between variables, using an empirical histogram to approximate the probability distribution yields better results. The part-based approach with the empirical distribution improved significantly the baseline detection that doesn't rely on tanks and piles part detectors. For that reason we use it in our large-scale experiments.
\input{data/val_acontrario_2}

\subsection{Impact of image resolution and source}
We conduct a study of the detection feasibility on several remote sensing sources with different levels of spatial resolution. To this end, we train our model using aerial BD ORTHO images at  0.5\,m and 1.5\,m respectively, SPOT (1.5m), and Sentinel-2 (10m). Results, shown in Table~\ref{tab:source_comparison}, indicate a significant performance drop with SPOT compared to aerial data at 0.5\,m, due to both resolution and domain shift. First, lower resolution reduces the apparent size of small targets such as tanks and piles. Second, annotations transferred from IGN data introduce temporal inconsistencies, unlabeled structures appear while others disappear. Nevertheless, SPOT results are close behind BD ORTHO at 1.5\,m/pixel. Since SPOT satellite data is more accessible and easier to process, it offers a reasonable trade-off between performance and efficiency. However, with \mbox{Sentinel-2}  on many objects detection fails entirely as they are depicted considerably small and/or blurry, largely underperforming compared to the rest of sources.

\begin{table}
\centering
\resizebox{\columnwidth}{!}{%
\begin{tabular}{@{}lllll|lll@{}}
\toprule
 & $mAP_{50}$ & Site & Pile & Tank & $AP_{dist}$ & Recall & Precision \\
\midrule
Aerial (0.5m) & \textbf{0.75} & 0.68 & \textbf{0.58} & \textbf{0.98} & \textbf{0.97} & 97.5\% & 97.5\% \\
Aerial (1.5m) & 0.61 & \textbf{0.76} & 0.17 & 0.89 & 0.93 & 95.0\% & 95.0\% \\
SPOT (1.5m) & 0.58 & 0.76 & 0.13 & 0.85 & 0.92 & 90.0\% & 97.3\% \\
Sentinel (10m) & 0.10 & 0.30 & 0.00 & 0.00 & 0.49 & 42.5\% & 85.0\% \\
\bottomrule
\end{tabular}
}
\caption{\textbf{Performance of the best object detection model across different image sources.} Recall and Precision are computed at the optimal F1 score using the distance-based matching.
}
\label{tab:source_comparison}
\end{table}

\subsection{Effect of the iterative process}
\input{imgs/imgs_hard_neg_main}
We evaluate the impact of the iterative process on the performance of our method and on the dataset as well. During the three iterations, we find 1) correct detections that were initially unlabeled and 2) challenging examples that were prone to be false positives. Table~\ref{tab:iterations_info} describes the introduction of these elements in each iteration. To this end, we collect a set of false positive examples that were not included during the dataset iterations, and include them in the validation set to evaluate the model in more challenging conditions. Some samples of these hard examples are shown in Figure~\ref{fig:hard_negatives_main}. We then evaluate our model in each iteration, as shown in Table~\ref{tab:hard_negative_ap50}. Performance improves with each iteration, with mAP50 going from 0.26 to 0.59. Tank detection improves the most, with AP50 rising from 0.16 to 0.86, hinting that the model learns to better separate bio-digester tanks from similar structures like oil tanks.


\begin{table}[t]
    \centering
    \small
    \begin{tabular}{lcccc}
        \toprule
        \textbf{Iteration} & \makecell{\textbf{Known} \\ \textbf{Database Size}} & \makecell{\textbf{New} \\ \textbf{Detection}} & \makecell{\textbf{Hard} \\ \textbf{Negatives}} \\
        \midrule
        0 & 203 & - & 0 \\
        1 & 203 & 149 & 100 \\
        2 & 352 & 205 & 200 \\
        \bottomrule
    \end{tabular}
    \caption{\textbf{Evolution of the dataset across iterations}. Summary of reviewed top-N detections, indicating counts of already known instances, new detections, and hard negatives for each iteration.
    }
    \label{tab:iterations_info}
\end{table}

\begin{table}[t]
\scriptsize
\centering
\resizebox{0.95\columnwidth}{!}{
  \begin{tabular}{@{}lrrrr|rr@{}}
    \toprule
    Iteration & $mAP_{50}$ & Site & Pile & Tank & \#Background & $\alpha$ \\
    \midrule
    0 & 0.26 & 0.52 & 0.11 & 0.14 & 163 & 50\% \\
    1 & 0.29 & 0.66 & 0.06 & 0.16 & 263 & 38\% \\
    2 & \textbf{0.59} & \textbf{0.76} & \textbf{0.14} & \textbf{0.86} & 363 & 31\% \\
    \bottomrule
  \end{tabular}
}
\caption{\textbf{Hard Negative Mining validation performance.}
}
\label{tab:hard_negative_ap50}
\end{table}

\subsection{Estimation of power production}
\begin{figure}
    \centering
    \includegraphics[width=\linewidth]{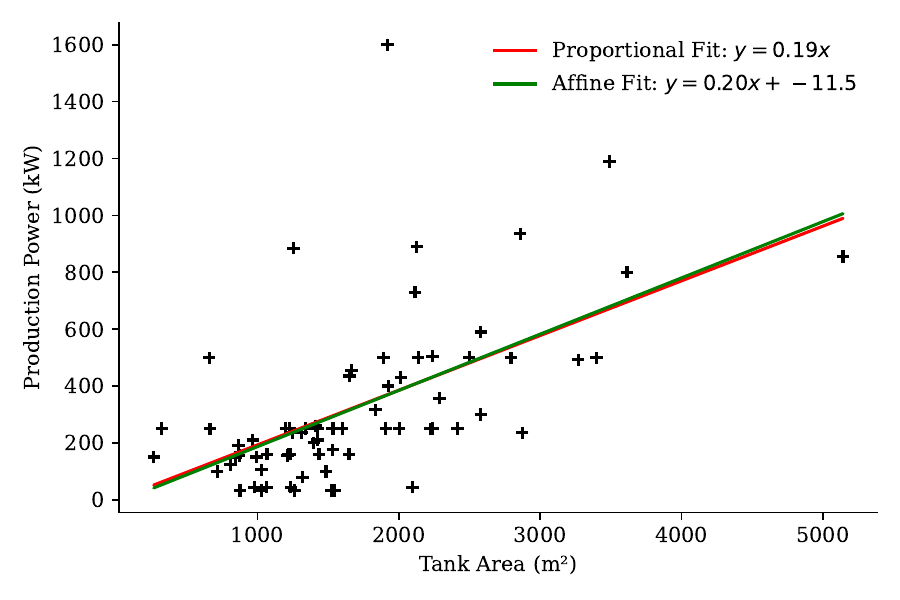}
    
    \vspace{-6pt}

    \caption{\textbf{Power Production (kW) against Tank Area ($m^2$) in Bretagne region ($r^2 = 0.332$)}.
    }
    \label{fig:Power_Prediction}
\end{figure}

Estimating power production of a large area can provide valuable information of the potential methane leaks and impact on the atmosphere. Thus, we leverage the generated bio-digester inventory of Bretagne to estimate its aggregated production in kilowatts (kW), for which we have the ground truth~\cite{AILE2025}. We build a linear regression model, illustrated in Figure~\ref{fig:Power_Prediction}, which predicts power production based on the overall detected tank area in a site (a feature linked to the scale of the facility). While unsurprisingly the model has a large predictive error, it is able to provide reasonable aggregated estimates over a large number of observations. Overall, the power production estimation model predicts 33\% of the power variability.

%% file: imgs/ap_grand_est.tex
\begin{figure}[t]
    \centering
    \includegraphics[trim={35 0 35 35},clip,width=0.5\textwidth]{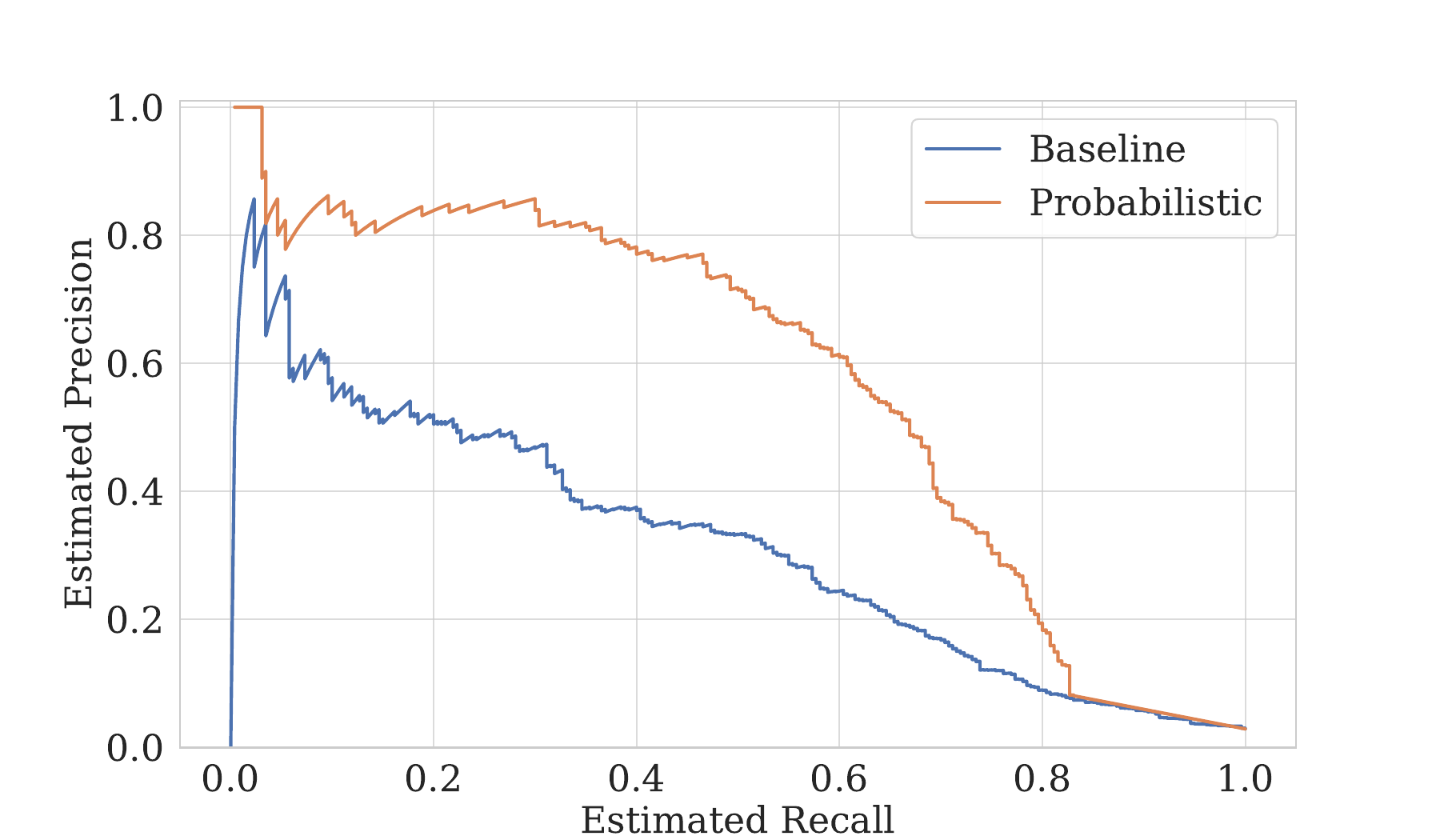} 
    \caption{\textbf{Precision-Recall curves for different detection methods} over all the Grand Est region. The Baseline method (blue) performs the worst, while the probabilistic part-based method (orange) significantly boosts the performance.
    }
    \label{fig:precision_grand_est}
\end{figure}

%% file: data/normandie_bars_2.tex
\begin{table}[t]
    \centering
    \small 
    \setlength{\tabcolsep}{4pt} 
    \begin{tabular}{l|ccc|ccc}
        \toprule
        \multirow{2}{*}{\textbf{Region}} & \multicolumn{3}{c|}{\textbf{In-Database}} & \multicolumn{3}{c}{\textbf{Overall}} \\
        \cmidrule(lr){2-4} \cmidrule(lr){5-7}
        & \textbf{TP} & \textbf{GT} & \textbf{Recall} & \textbf{Correct} & \textbf{Total} & \textbf{Precision} \\
        \midrule
        Grand Est & 188 & 224  & 83.9\% & 251 & 311 & 80.7\% \\
        Marne     & 18  & 21   & 85.7\% & 28  & 36  & 77.8\%  \\
        Bretagne  & 96  & 179  & 53.6\% & 131 & 185 & 70.8\% \\
        \bottomrule
    \end{tabular}
    \caption{\textbf{Detection results in Marne, Grand Est, and Bretagne sub-regions}. Results are split between in-database (existing ground truth) and overall detections, including new discoveries. Precision and recall are computed using confirmed matches; overall counts include detections outside the reference database.}
    \label{tab:detection_results}
\end{table}


%% file: data/val_acontrario_2.tex
\setlength{\tabcolsep}{4pt}

\begin{table}[t]
\centering
\small
  \begin{tabular}{@{}lccc@{}}
    \toprule
Method & Probability  & $AP_{dist}$ & Max Recall at \\
       &     Law      &  & 100\% Precision \\
    \specialrule{1.2pt}{1pt}{2pt}
Baseline &  & 0.949 & 62.5 \% \\
\midrule
Part-based & Poisson & 0.917 & 62.5 \% \\
Part-based & Histogram & \textbf{0.969} & \textbf{95.0 \%} \\

    \bottomrule
  \end{tabular}
  \caption{\textbf{Ablation of proposed approaches to reduce false positives}. We report their impact based on average precision ($AP_{dist}$) and the maximum recall at 100\% precision.}
  \label{tab:fp_ablation}
\end{table}

%% file: imgs/imgs_hard_neg_main.tex
\begin{figure}[t]
  \centering
  \begin{subfigure}[t]{0.32\columnwidth}
    \centering
    \includegraphics[width=\linewidth,trim=300 300 300 300,clip]{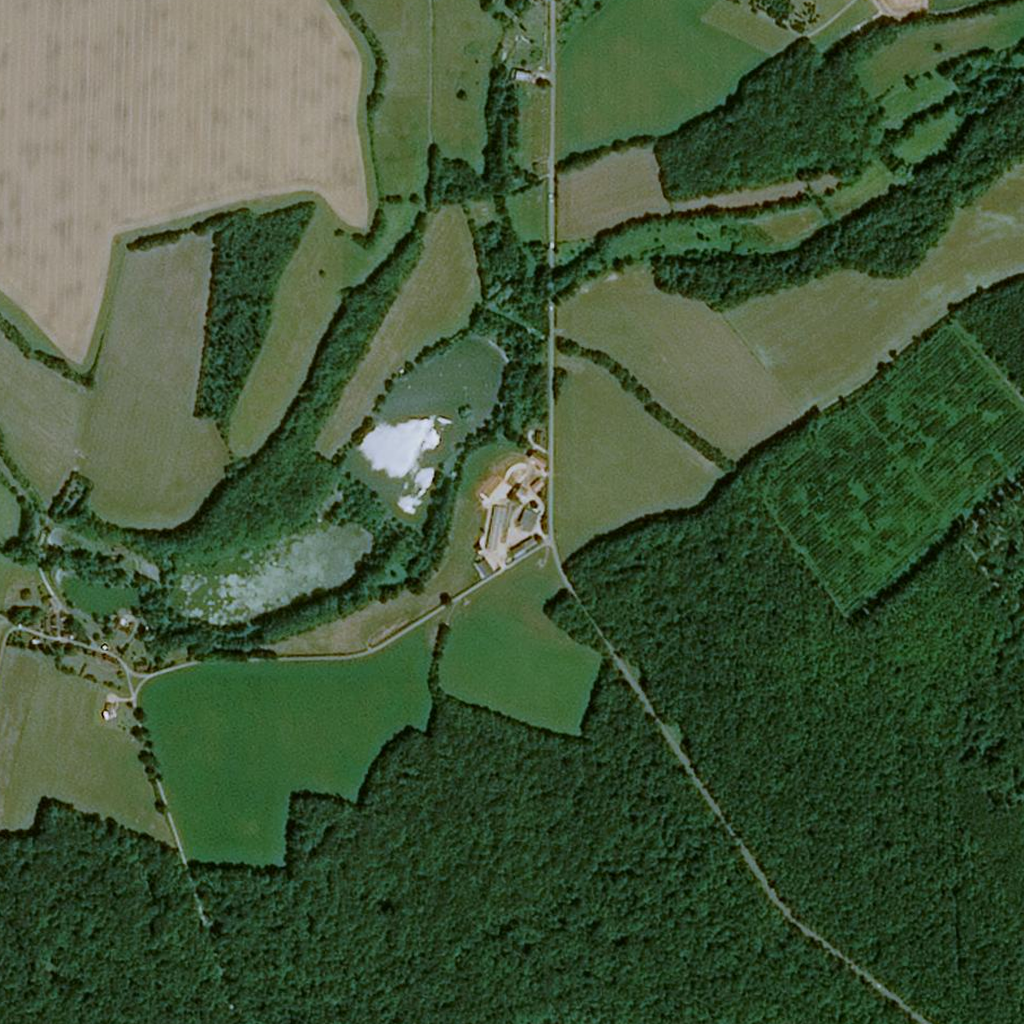}
    \label{fig:hn_main_iter0}
  \end{subfigure}\hfill
  \begin{subfigure}[t]{0.32\columnwidth}
    \centering
    \includegraphics[width=\linewidth,trim=300 300 300 300,clip]{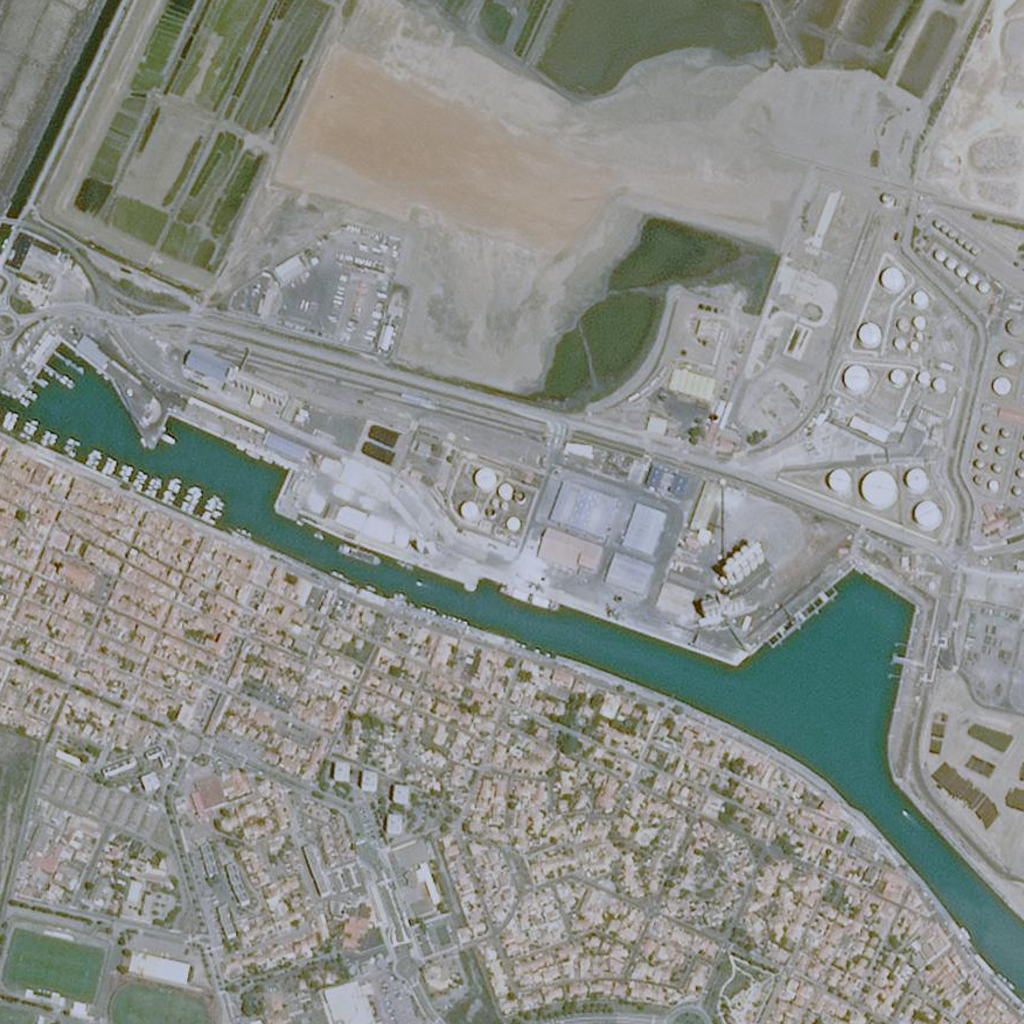}
    \label{fig:hn_main_iter1}
  \end{subfigure}\hfill
  \begin{subfigure}[t]{0.32\columnwidth}
    \centering
    \includegraphics[width=\linewidth,trim=300 300 300 300,clip]{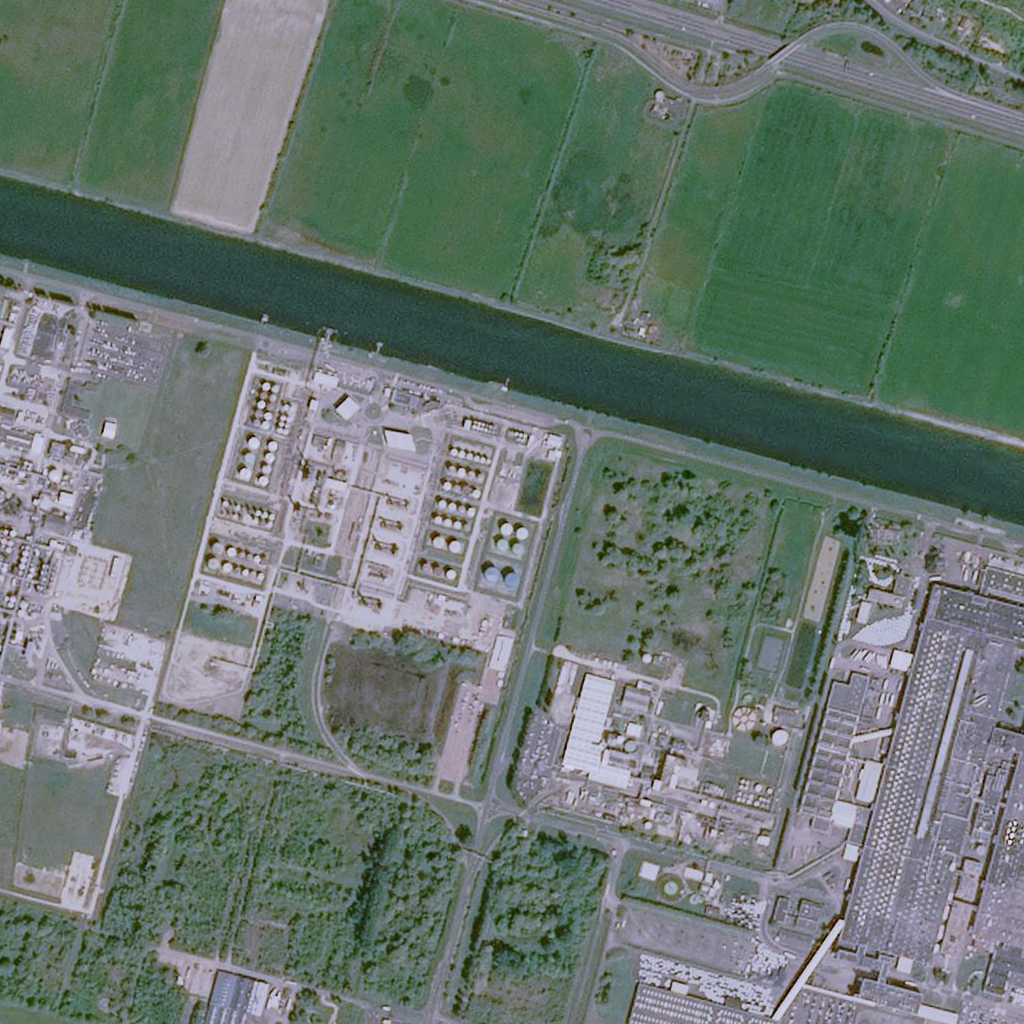}
    \label{fig:hn_main_iter2}
  \end{subfigure}
  \vspace{-10pt}
  
  \caption{\textbf{Examples of hard negatives} found during the iterative process. They were initially detected as false positives, as they do not contain bio-digester sites.\vspace{-6pt}}
  \label{fig:hard_negatives_main}
\end{figure}

%% file: sec/4_end.tex
\section{Conclusion}
In this work, we propose a method to detect bio-digester sites from limited and incomplete annotations, combining part-based detection and hard negative mining to counter the sheer amount of false-positives classical detectors produces at large scale.
We demonstrate how our detector can learn to detect bio-digester sites robustly within the training area (Grand Est) are and generalize to other regions as well (Bretagne). The proposed iterative process allowed us to find new, unlabeled sites and false positive sensitive observations, all which we used to complement the databases and build a large-scale dataset. Furthermore, we provide several ablations and evaluations of our method and its components, as well as a study of the ideal resolution for an optimal trade-off between performance and large-scale efficiency. Lastly, we built a regression model that can provide a reasonable estimate of the power production of an entire area. We release our code, data, and detected locations to support future research. 

\paragraph{Future work.} This article introduces the idea of monitoring methane-contributions from small emitters through bottom-up inventorying approaches. Whilst it proves effective, future works can expand this in several directions. First, this study opens the possibility to apply similar procedures to country-level areas, computing potential leaks of all of France or any other state. Moreover, SPOT satellite feasibility enables temporal analyses of bio-digester construction and, therefore, yearly growth of methane production. We believe that this work lays the groundwork for broader small emitter methane mapping and the development of more robust detectors, thus supporting the Earth observation community.

%% file: sec/X_suppl.tex
\clearpage
\setcounter{page}{1}
\maketitlesupplementary

\subsection{Benchmarking Object Detectors}
We compare and justify our object detection architecture choice using the annotated validation data. 
We evaluate several object detection architectures to determine the most suitable for our task. We consider YOLOv8~\cite{yolov8} for its efficiency, Faster R-CNN~\cite{fasterrcnn} as a lightweight two-stage detector, and two oriented object detectors: Oriented R-CNN~\cite{Oriented_RCNN} and LSKNet~\cite{li2023largeselectivekernelnetwork}. Table~\ref{tab:ap50_scores} summarizes the overall and per-class AP50 scores for all models using BD ORTHO images at 1m per pixel resolution.

We first assess the impact of pretraining by comparing the fine-tuned performance of YOLOv8 models pretrained on DOTA~\cite{dota} and COCO~\cite{COCO}. The results confirm that domain-specific pretraining with DOTA improves YOLOv8 performance by 74\%.
Oriented object detectors not only enhance pile detection but also improve overall bio-digester identification. However, for tank detection, traditional object detectors perform better. This is not surprising, as rounded objects lack a clear orientation, a known problem in oriented object detection~\cite{structure_tensor_obb}.
Based on this and the above results, we select the LSKNet as our final object detector. Figure~\ref{fig:maps_grand_est} (b) presents qualitative examples from the validation set.

\AD{}{

\input{data/val_models_2}

}

\subsection{Oriented Bounding Boxes}
\input{imgs/piles_inclined}
Figure~\ref{fig:imgs_piles_inclined} illustrates a case where oriented bounding boxes provide a more suitable annotation format. Piles are typically slender and not aligned with the horizontal or vertical axes, leading axis-aligned bounding boxes to produce artificially high Intersection-over-Union (IoU) scores due to excessive overlap.

\subsection{Training Implementation Details}
\paragraph{Faster RCNN} Uses Detectron~\cite{wu2019detectron2} implementation. Is trained using Smooth L1 loss for bounding box regression and Focal Loss for classification ($\alpha=0.25, \gamma=2.0$). Regularization includes a weight decay of 0.0001, while batch normalization is frozen. No dropout is applied.
\paragraph{YOLOv8} Uses mmyolo~\cite{mmyolo2022} implementation. It Is trained with a classification loss weight of 0.5, a bounding box regression loss weight of 7.5, and a distribution focal loss weight of 0.375. The base learning rate is set to 0.01 with a linear warmup schedule and learning rate decay. uses batch normalization with a momentum of 0.03. The model applies L2 regularization with a weight decay of 0.0001. Batch normalization remains frozen. No dropout is applied.

\paragraph{Oriented RCNN and LSKNet} Use mmrotate~\cite{zhou2022mmrotate} implementation. The RPN is trained with a head uses a sigmoid-based Cross Entropy loss (weight = 1.0) for objectness classification and a Smooth L1 loss (beta = 0.111) for bounding box regression. The RoI head applies a softmax-based Cross Entropy loss (weight = 1.0) for multi-class classification and a Smooth L1 loss (beta = 1.0, weight = 1.0) for final box refinement. 
The initial learning rate is set to 0.01 and follows a linear warm-up phase, starting at 0.001 and increasing over 500 iterations. After the warm-up, a MultiStepLR scheduler is applied, reducing the learning rate by a factor of 0.1 at epoch 7. The training process is conducted over a total of 8 epochs. Models incorporates L2 regularization with a weight decay of 0.0001. Batch normalization layers are frozen during training. No dropout is applied.

\input{imgs/imgs_hard_neg_examples}


%% file: data/val_models_2.tex
\setlength{\tabcolsep}{7pt}
\begin{table}[t]
\resizebox{\columnwidth}{!}{
\centering
  \begin{tabular}{@{}lcccccc@{}}
    \toprule
Model  & Overall & Bio-digester & Tank & Pile \\
    \midrule
Yolov8 (COCO) \cite{yolov8} & 0.386 & 0.383 & 0.601 & 0.173 \\
Yolov8 (DOTA) \cite{yolov8}  & 0.672 & 0.736 & 0.901 & 0.378 \\
Faster RCNN \cite{ren2016fasterrcnnrealtimeobject} & 0.687 & \textbf{0.854} & \textbf{0.909} & 0.299 \\
\midrule
LSK \cite{li2023largeselectivekernelnetwork} & \textbf{0.724} & 0.846 & 0.741 & \textbf{0.584} \\
Oriented RCNN \cite{Oriented_RCNN} &  0.701 & 0.824 & 0.746 & 0.533 \\

    \bottomrule
  \end{tabular}
  }
  \caption{\textbf{Comparison of different object detectors}. Overall and per-class AP50 scores are reported for different models and parameters. YOLOv8 (COCO) and YOLOv8 (DOTA) correspond to the YOLO model pre-trained on COCO and DOTA, respectively.}
  \label{tab:ap50_scores}

\end{table}

%% file: imgs/piles_inclined.tex
\begin{figure}[ht]
    \centering
    \begin{subfigure}{0.48\linewidth}
        \centering
        \includegraphics[width=\linewidth]{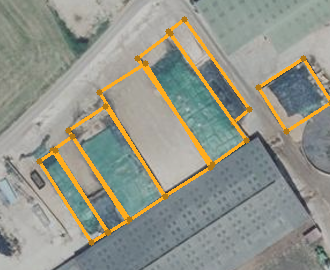}
        \caption{Oriented bounding boxes without overlap}
        \label{fig:imgs_piles_inclined_obb}
    \end{subfigure}
    \hfill
    \begin{subfigure}{0.48\linewidth}
        \centering
        \includegraphics[width=\linewidth]{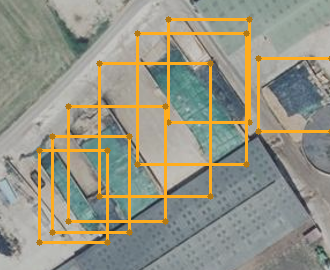}
        \caption{High overlap failure case of bounding boxes}
        \label{fig:imgs_piles_inclined_bbox}
    \end{subfigure}
    \caption{Visualization of annotation methods on piles}
    \label{fig:imgs_piles_inclined}
\end{figure}

%% file: imgs/imgs_hard_neg_examples.tex
\newcommand{\figwidth}{0.19\textwidth}       
\newcommand{\rowgap}{0.5em}                  

\begin{figure*}[ht]
  \centering
  \setlength{\tabcolsep}{1pt}
  \begin{tabular}{@{}p{0.5cm}@{}ccccc@{}} 
    & \multicolumn{5}{c}{\textbf{Hard Negative Examples}} \\[\rowgap]

    \rotatebox{90}{\hspace{23pt}\textbf{Iteration 0}} &
    \includegraphics[width=\figwidth,trim=300 300 300 300,clip]{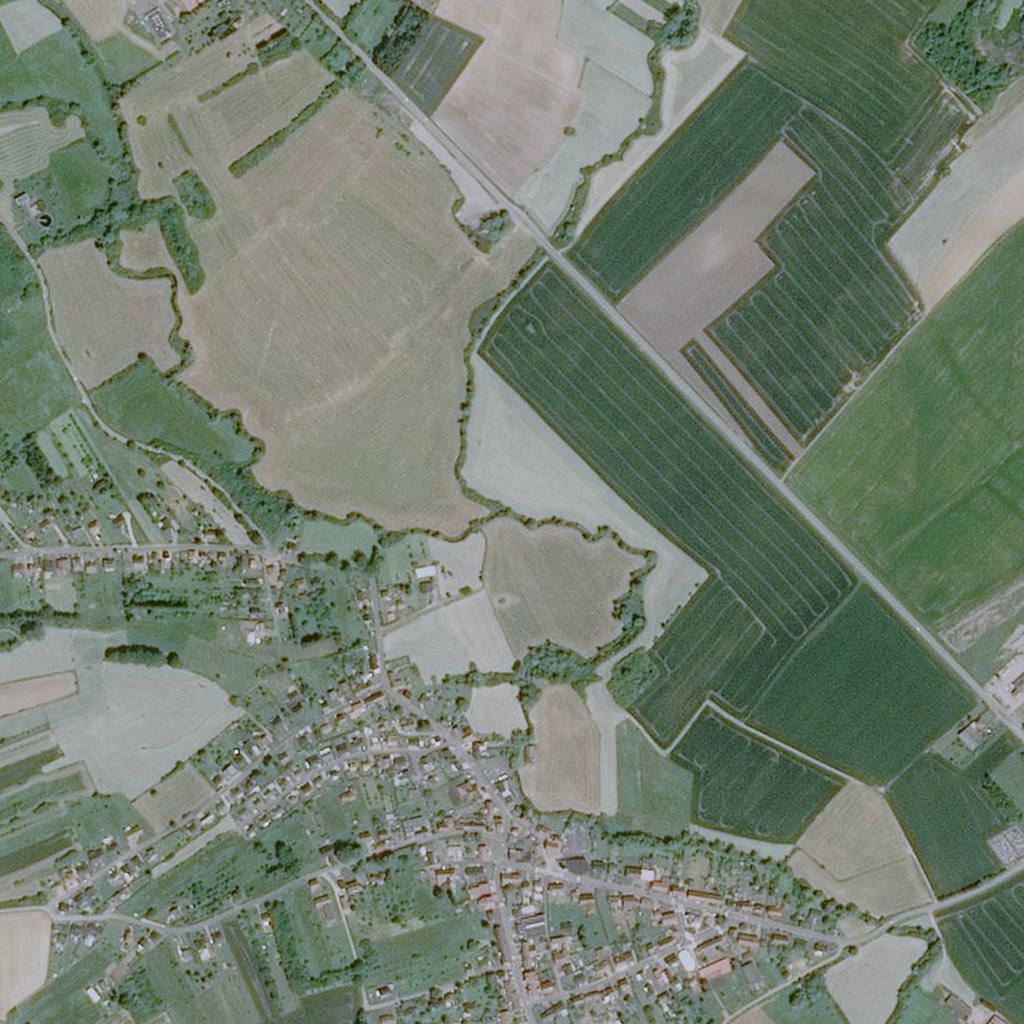} &
    \includegraphics[width=\figwidth,trim=300 300 300 300,clip]{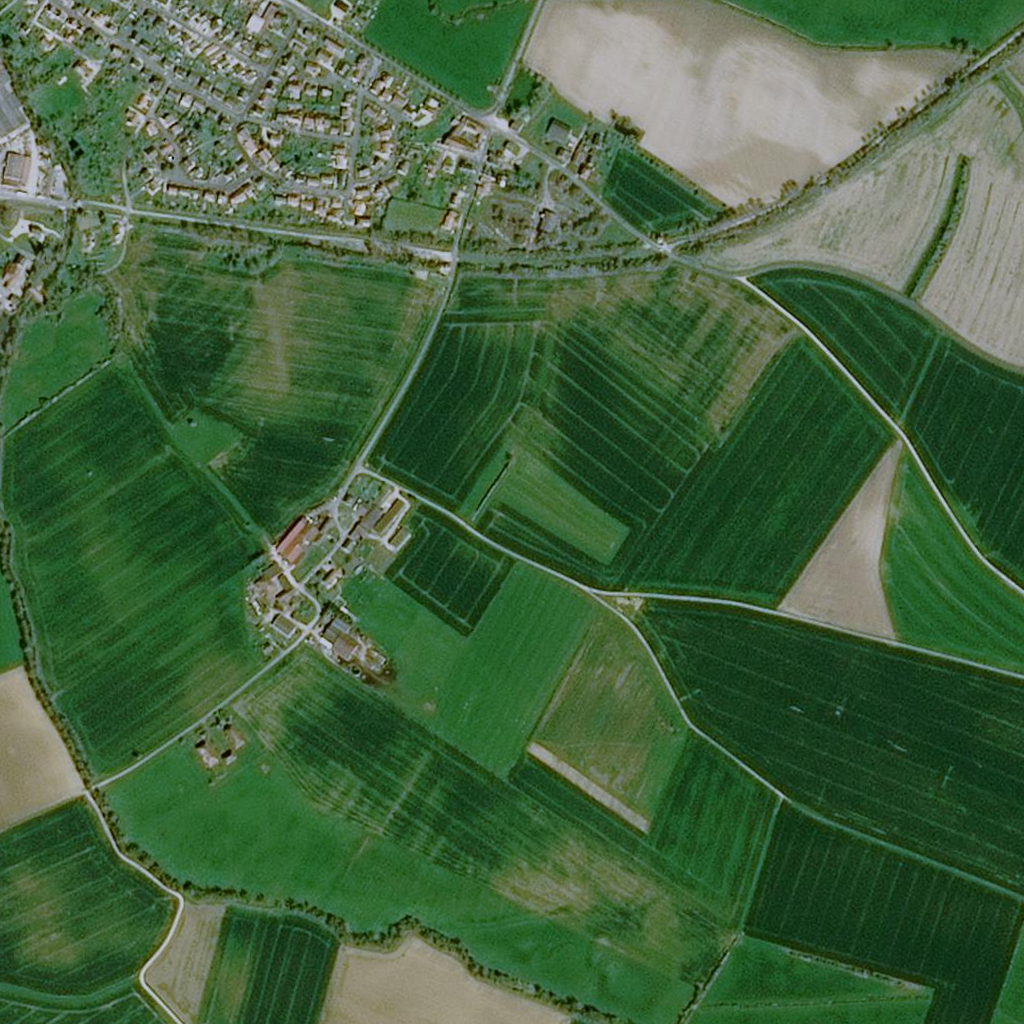} &
    \includegraphics[width=\figwidth,trim=300 300 300 300,clip]{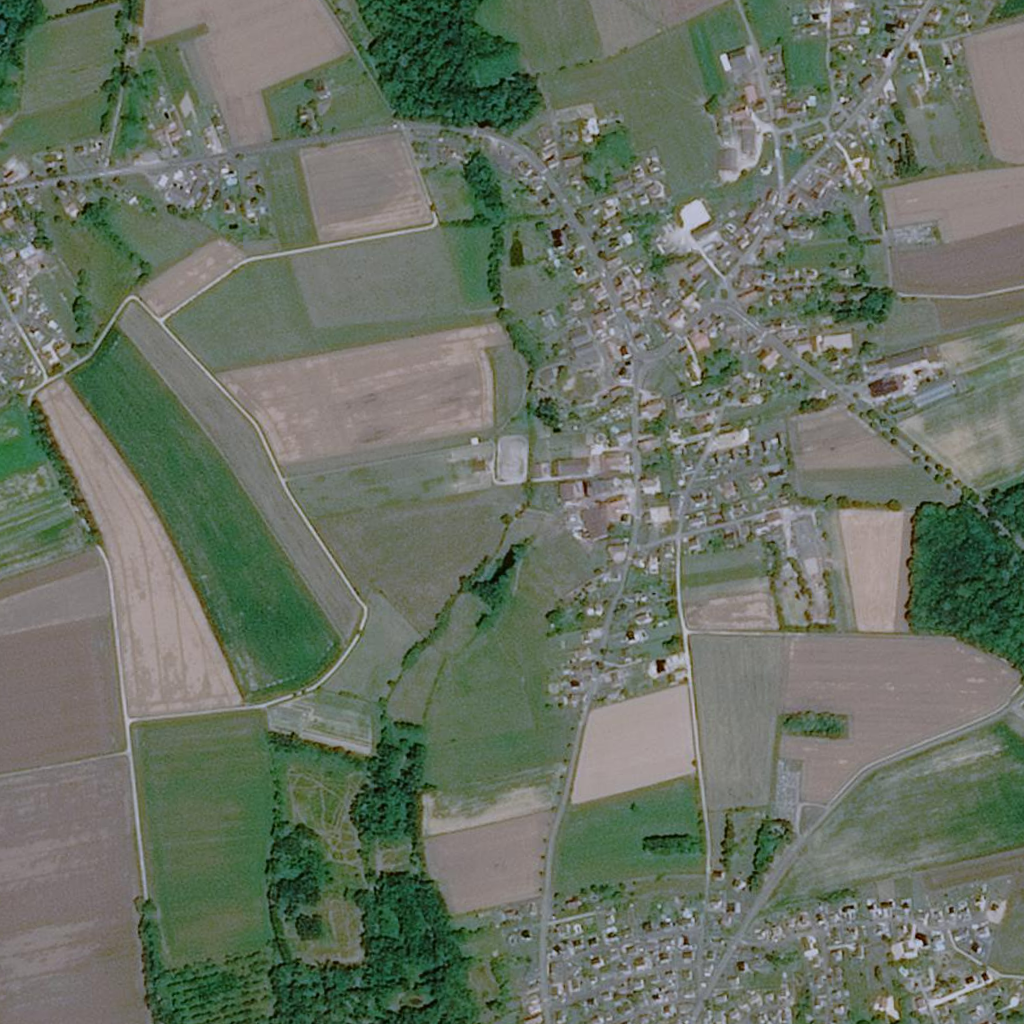} &
    \includegraphics[width=\figwidth,trim=300 300 300 300,clip]{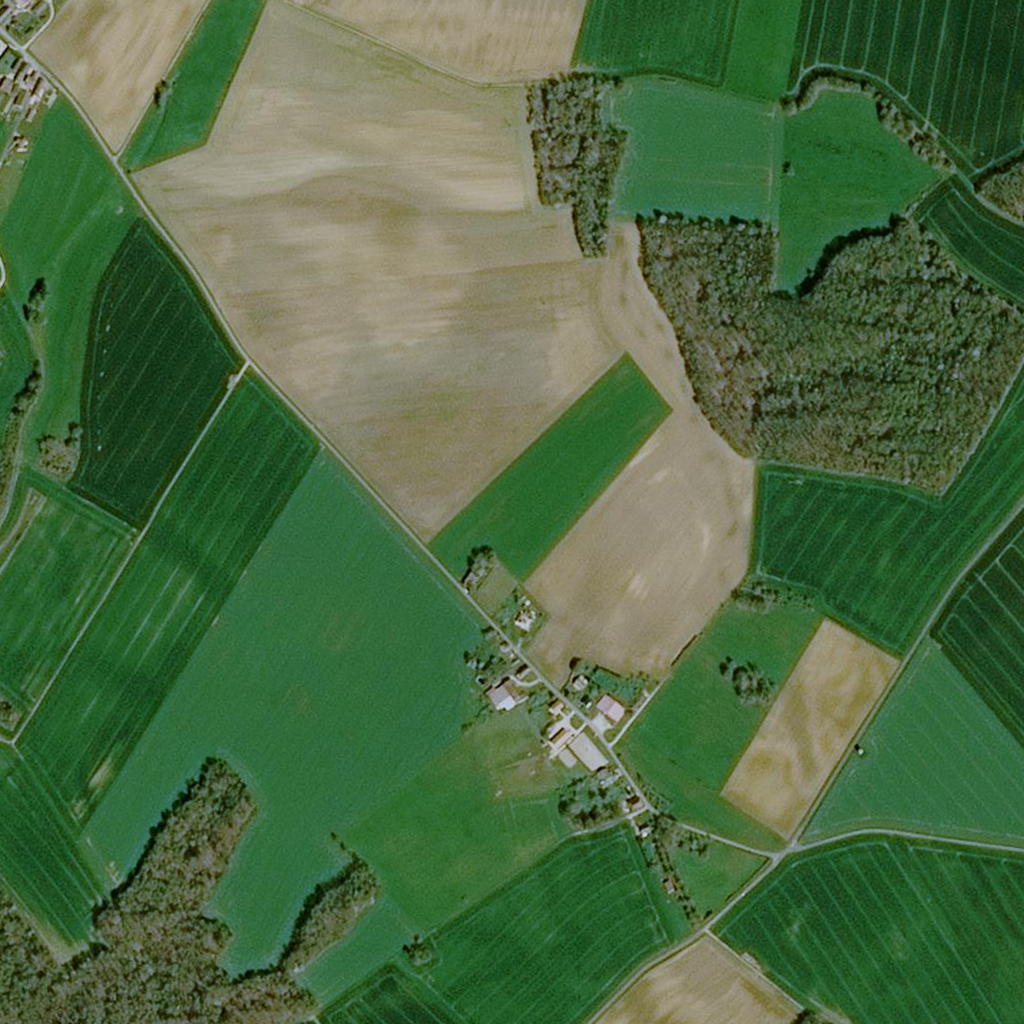} &
    \includegraphics[width=\figwidth,trim=300 300 300 300,clip]{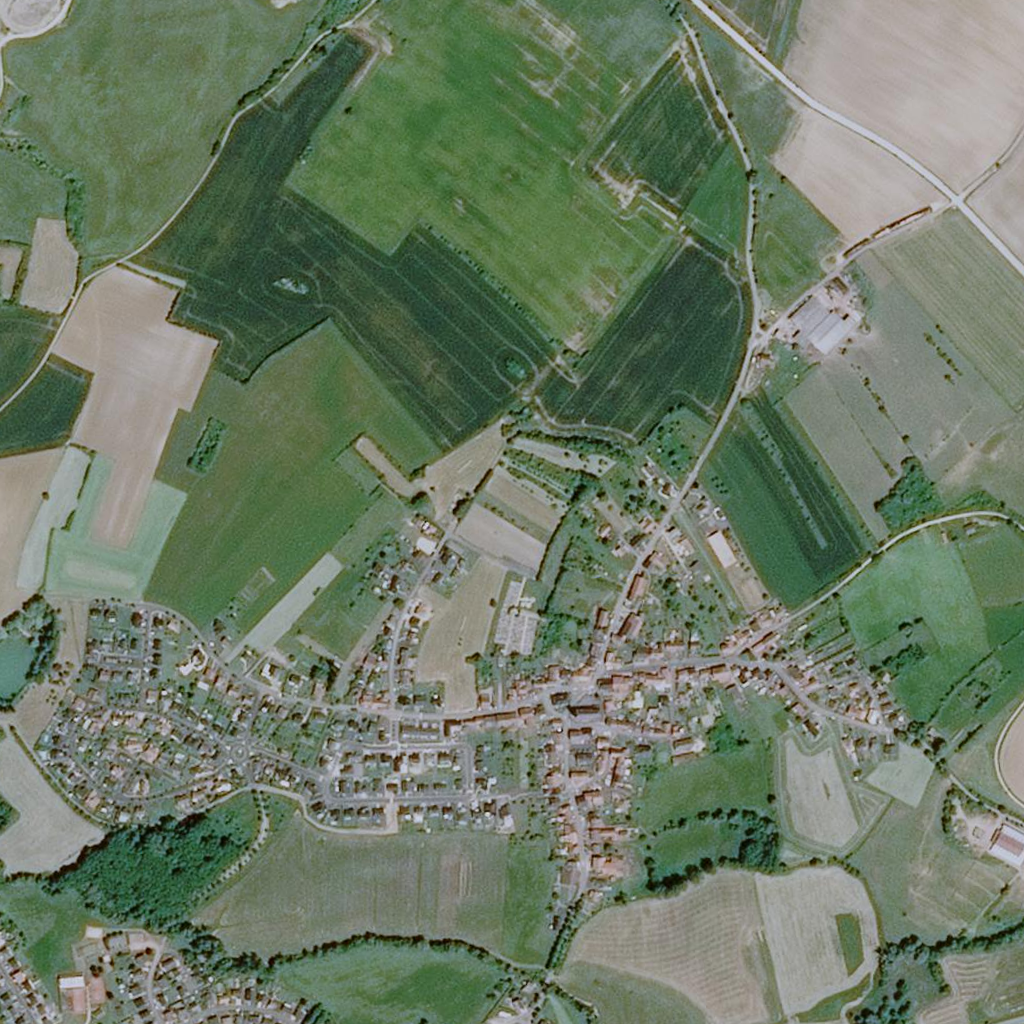}
    \\[\rowgap]

    \rotatebox{90}{\hspace{23pt}\textbf{Iteration 1}} &
    \includegraphics[width=\figwidth,trim=300 300 300 300,clip]{imgs/hard_negatives_examples/iter_01_False_2592.png} &
    \includegraphics[width=\figwidth,trim=300 300 300 300,clip]{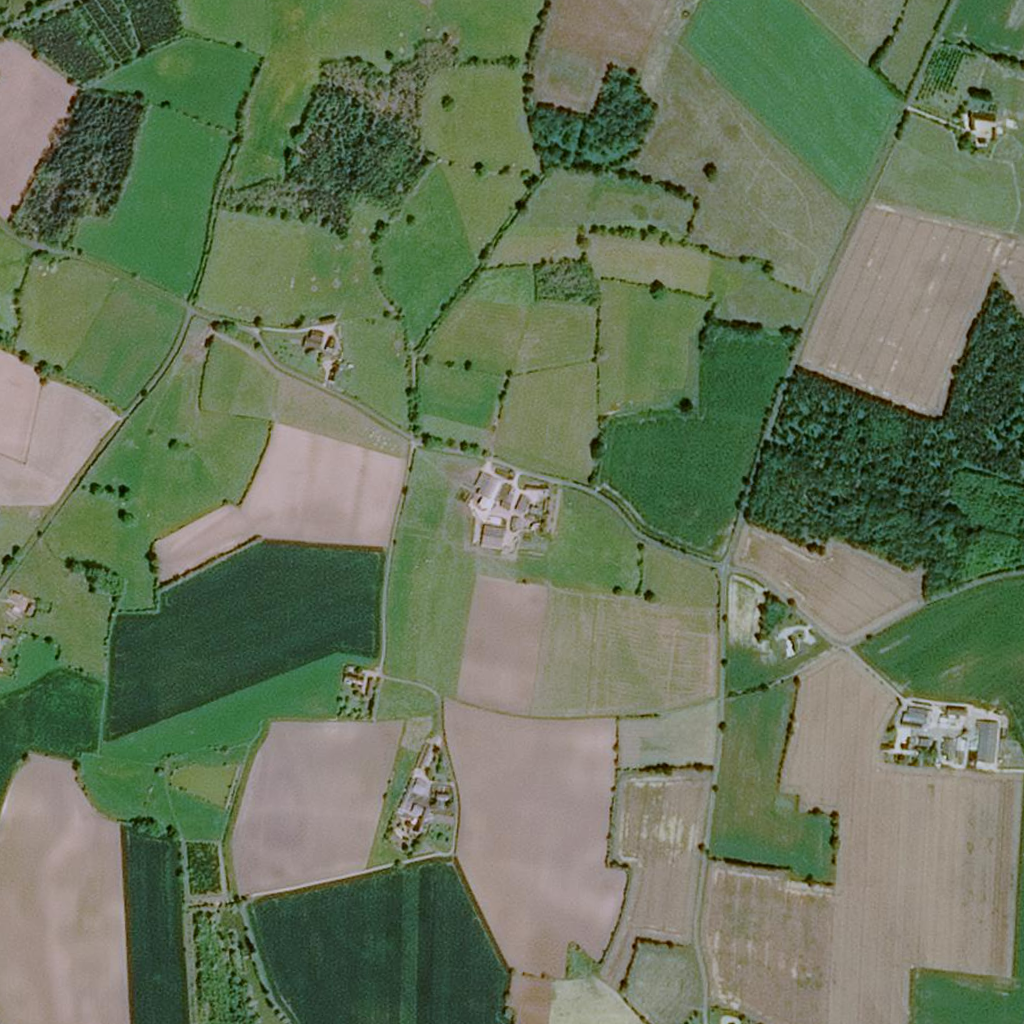} &
    \includegraphics[width=\figwidth,trim=300 300 300 300,clip]{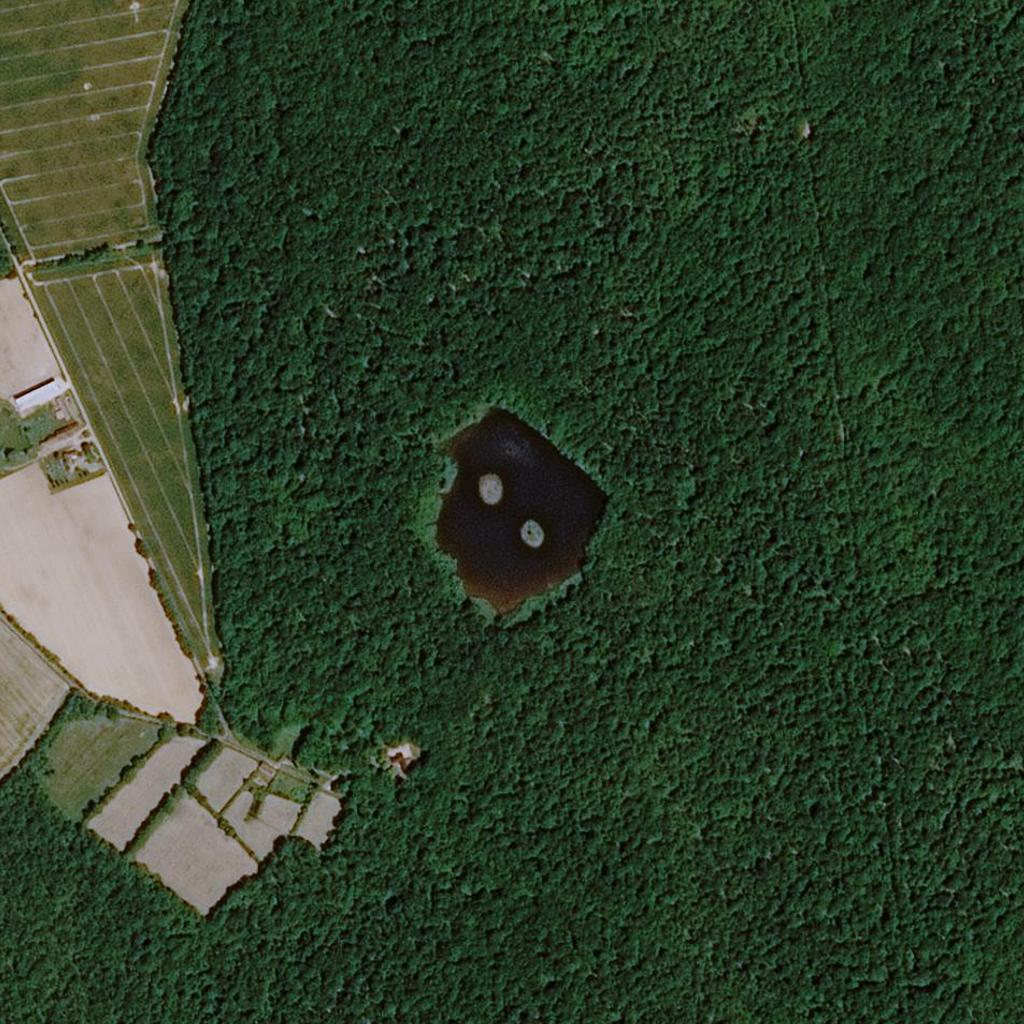} &
    \includegraphics[width=\figwidth,trim=300 300 300 300,clip]{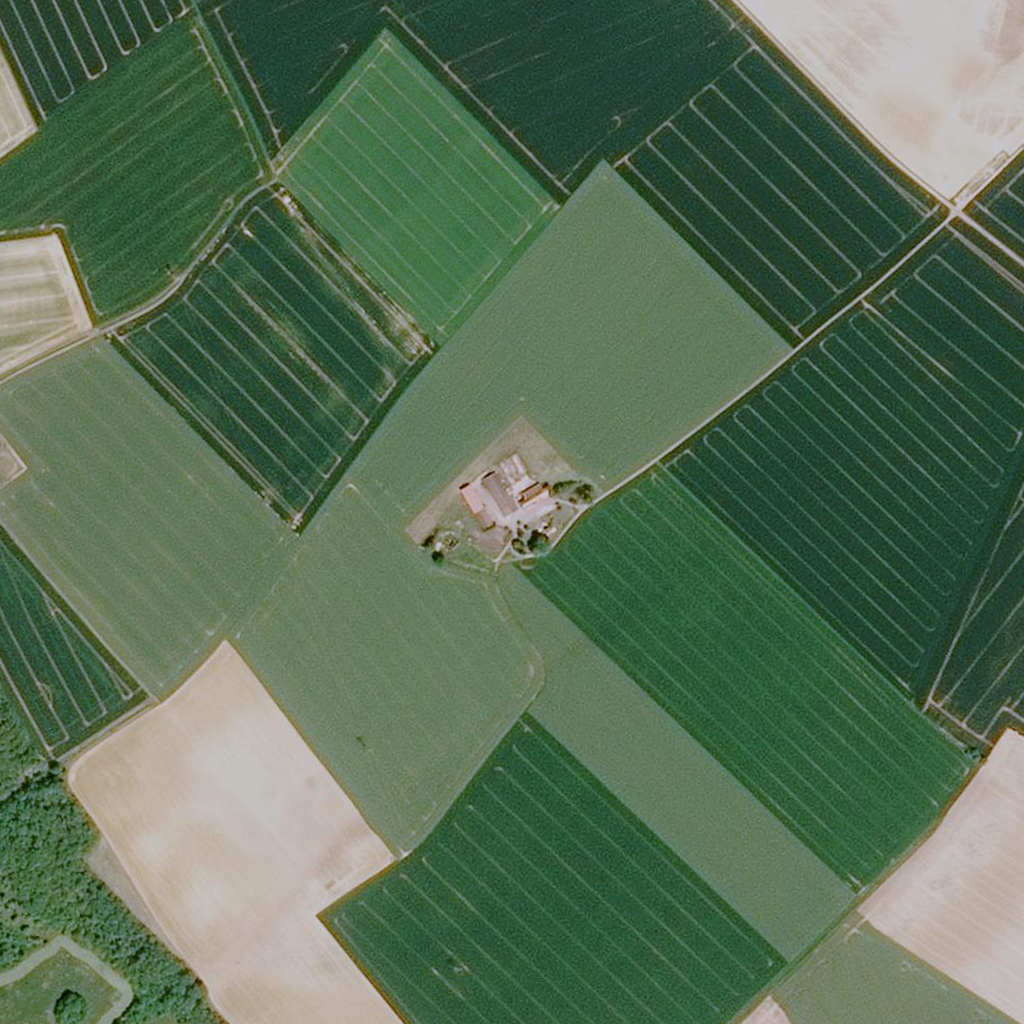} &
    \includegraphics[width=\figwidth,trim=300 300 300 300,clip]{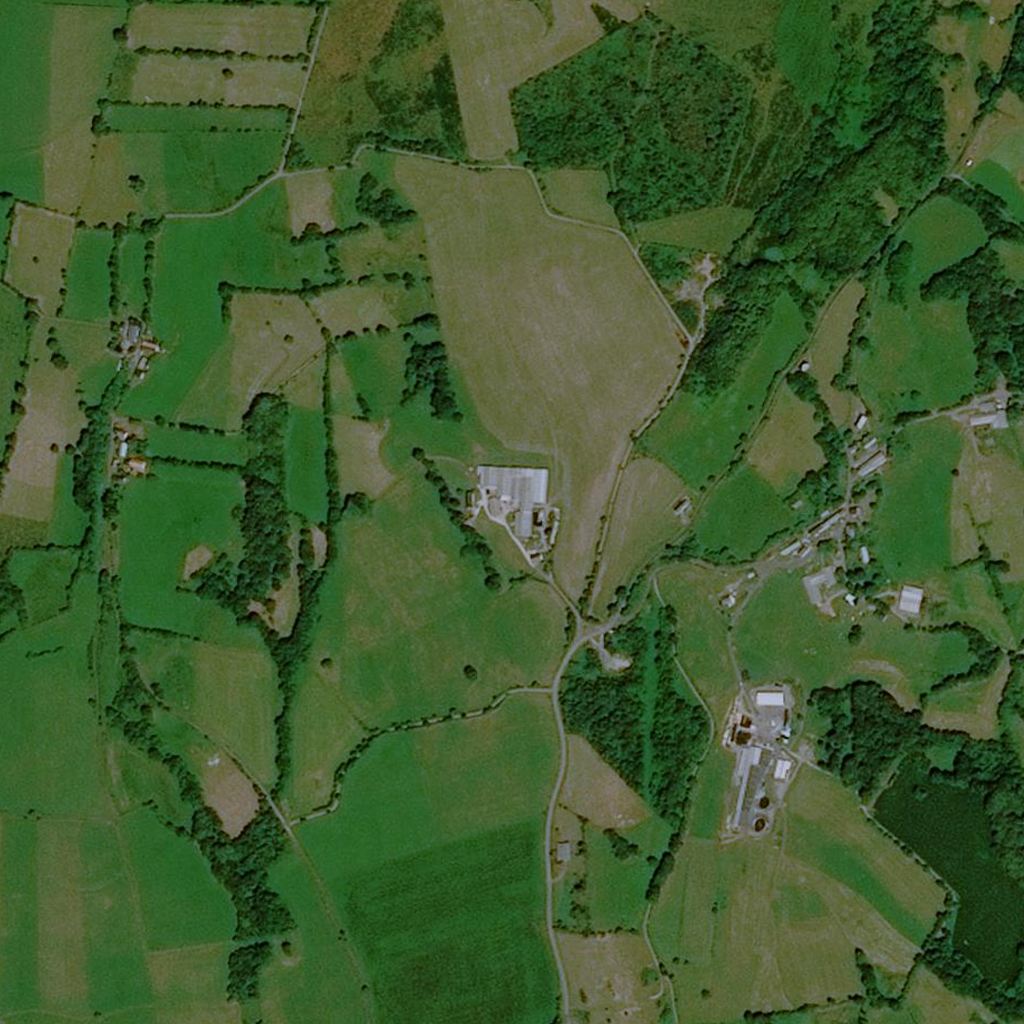}
    \\[\rowgap]

    \rotatebox{90}{\hspace{23pt}\textbf{Iteration 2}} &
    \includegraphics[width=\figwidth,trim=300 300 300 300,clip]{imgs/hard_negatives_examples/iter_02_False_4307.png} &
    \includegraphics[width=\figwidth,trim=300 300 300 300,clip]{imgs/hard_negatives_examples/iter_02_False_4304.png} &
    \includegraphics[width=\figwidth,trim=300 300 300 300,clip]{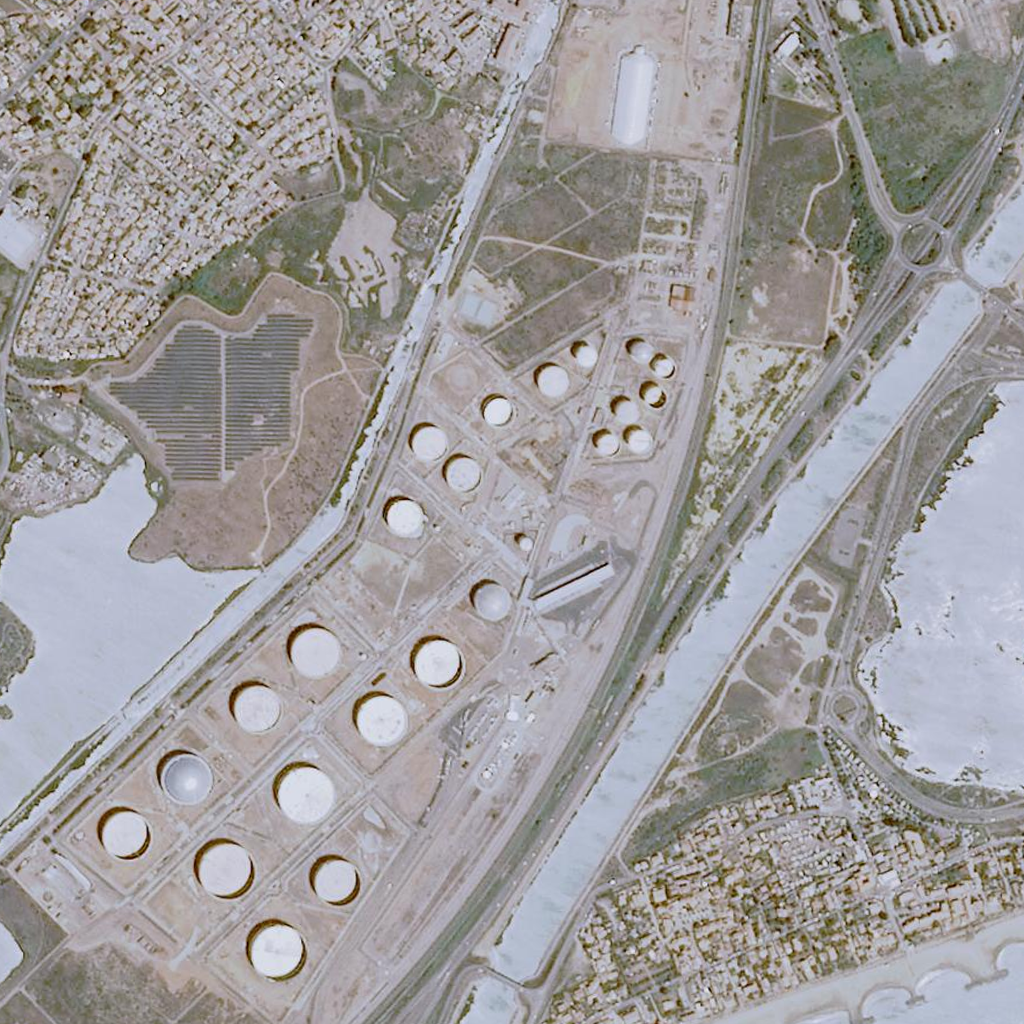} &
    \includegraphics[width=\figwidth,trim=300 300 300 300,clip]{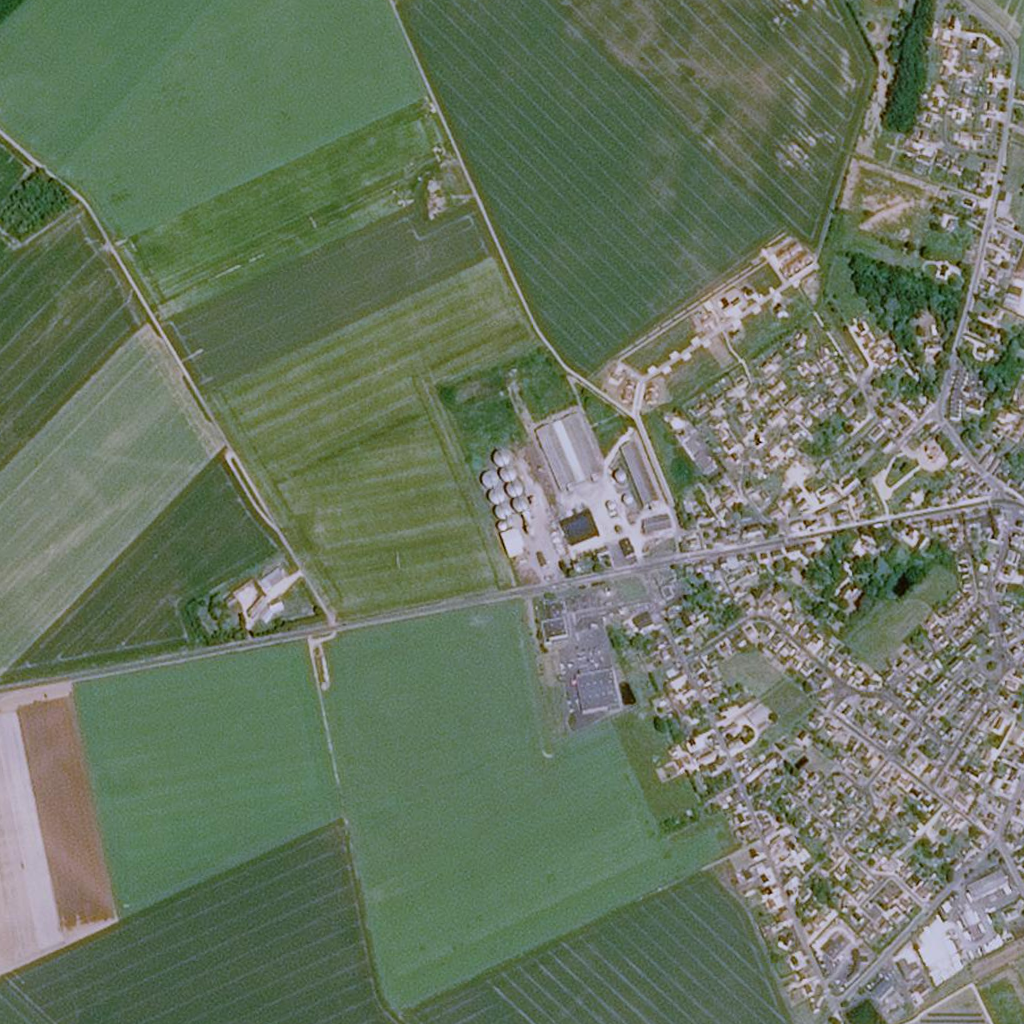} &
    \includegraphics[width=\figwidth,trim=300 300 300 300,clip]{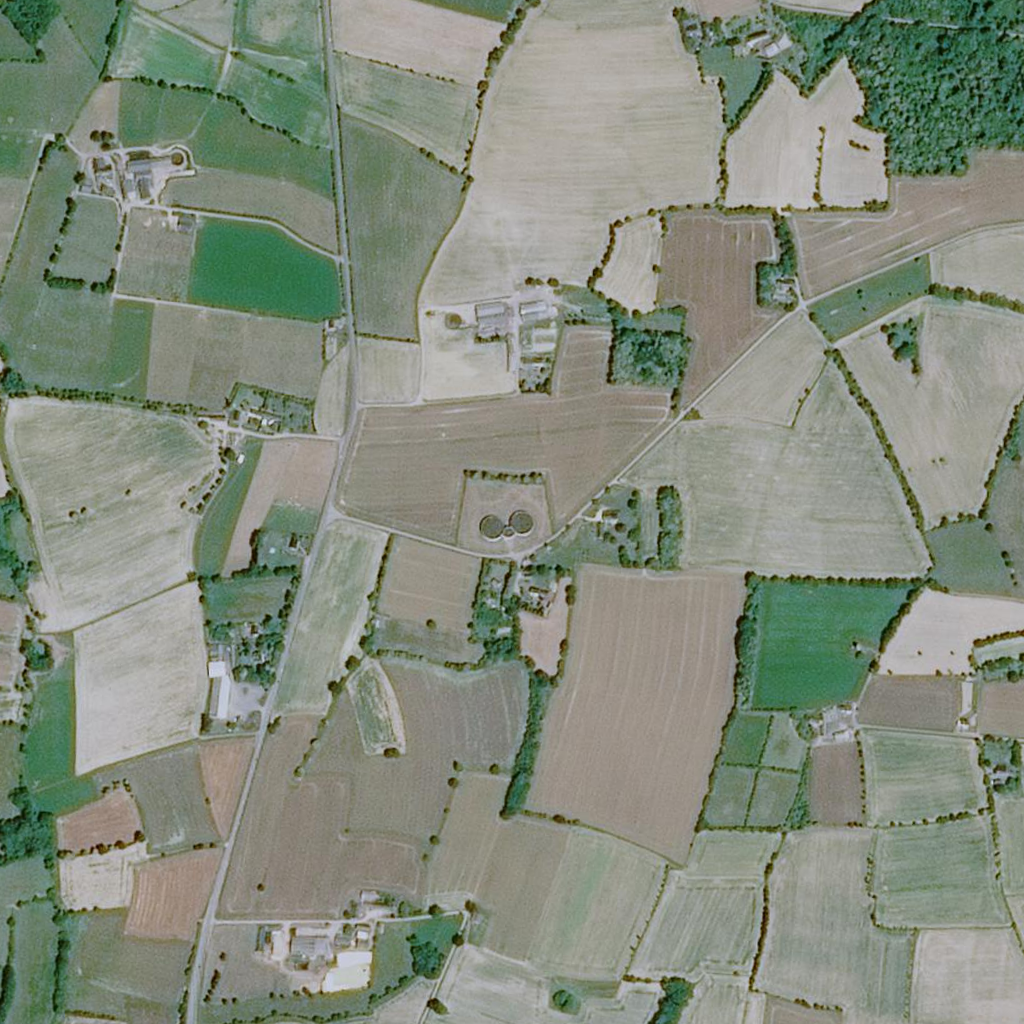}
  \end{tabular}

   \caption{\textbf{Hard negative examples across three training iterations.}
    At iteration 0, samples are randomly drawn from the Grand-Est region.}
  \label{fig:imgs_hard_negatives}
\end{figure*}

%% file: main.bbl
\begin{thebibliography}{75}
\providecommand{\natexlab}[1]{#1}
\providecommand{\url}[1]{\texttt{#1}}
\expandafter\ifx\csname urlstyle\endcsname\relax
  \providecommand{\doi}[1]{doi: #1}\else
  \providecommand{\doi}{doi: \begingroup \urlstyle{rm}\Url}\fi

\bibitem[{Association AILE}(2025)]{AILE2025}
{Association AILE}.
\newblock Carte \& chiffres clés - la filière méthanisation en bretagne et pays de la loire.
\newblock \url{https://aile.asso.fr/biogaz/la-filiere/carte-chiffres-cles-methanisation/}, 2025.
\newblock Accessed: June 12, 2025.

\bibitem[Bou et~al.(2023)Bou, Artola, Ehret, Facciolo, Morel, and von Gioi]{bou2023}
Xavier Bou, Aitor Artola, Thibaud Ehret, Gabriele Facciolo, Jean-Michel Morel, and Rafael~Grompone von Gioi.
\newblock Reducing false alarms in video surveillance by deep feature statistical modeling, 2023.

\bibitem[Bou et~al.(2024{\natexlab{a}})Bou, Facciolo, von Gioi, Morel, and Ehret]{bou2024exploringrobustfeaturesfewshot}
Xavier Bou, Gabriele Facciolo, Rafael~Grompone von Gioi, Jean-Michel Morel, and Thibaud Ehret.
\newblock Exploring robust features for few-shot object detection in satellite imagery, 2024{\natexlab{a}}.

\bibitem[Bou et~al.(2024{\natexlab{b}})Bou, Facciolo, von Gioi, Morel, and Ehret]{structure_tensor_obb}
Xavier Bou, Gabriele Facciolo, Rafael~Grompone von Gioi, Jean-Michel Morel, and Thibaud Ehret.
\newblock Structure tensor representation for robust oriented object detection, 2024{\natexlab{b}}.

\bibitem[Cheng and Han(2016)]{obj_det_survey_remote_sensing}
Gong Cheng and Junwei Han.
\newblock A survey on object detection in optical remote sensing images.
\newblock \emph{ISPRS journal of photogrammetry and remote sensing}, 117:\penalty0 11--28, 2016.

\bibitem[Cheng et~al.(2016)Cheng, Zhou, and Han]{obj_rs_3}
Gong Cheng, Peicheng Zhou, and Junwei Han.
\newblock Learning rotation-invariant convolutional neural networks for object detection in vhr optical remote sensing images.
\newblock \emph{IEEE Transactions on Geoscience and Remote Sensing}, 54\penalty0 (12):\penalty0 7405--7415, 2016.

\bibitem[Contributors(2022)]{mmyolo2022}
MMYOLO Contributors.
\newblock {MMYOLO: OpenMMLab YOLO} series toolbox and benchmark.
\newblock \url{https://github.com/open-mmlab/mmyolo}, 2022.

\bibitem[Cusworth et~al.(2022)Cusworth, Thorpe, Ayasse, Stepp, Heckler, Asner, Miller, Yadav, Chapman, Eastwood, et~al.]{cusworth2022strong}
Daniel~H. Cusworth, Andrew~K. Thorpe, Alana~K. Ayasse, David Stepp, Joseph Heckler, Gregory~P. Asner, Charles~E. Miller, Vineet Yadav, John~W. Chapman, Michael~L. Eastwood, et~al.
\newblock Strong methane point sources contribute a disproportionate fraction of total emissions across multiple basins in the united states.
\newblock \emph{Proceedings of the National Academy of Sciences}, 119\penalty0 (38):\penalty0 e2202338119, 2022.

\bibitem[Ding et~al.(2019)Ding, Xue, Long, Xia, and Lu]{rot_transf}
Jian Ding, Nan Xue, Yang Long, Gui-Song Xia, and Qikai Lu.
\newblock Learning roi transformer for oriented object detection in aerial images.
\newblock In \emph{Proceedings of the IEEE/CVF conference on computer vision and pattern recognition}, pages 2849--2858, 2019.

\bibitem[Duan et~al.(2019)Duan, Bai, Xie, Qi, Huang, and Tian]{centernet}
Kaiwen Duan, Song Bai, Lingxi Xie, Honggang Qi, Qingming Huang, and Qi Tian.
\newblock Centernet: Keypoint triplets for object detection.
\newblock In \emph{2019 IEEE/CVF International Conference on Computer Vision (ICCV)}, pages 6568--6577, 2019.

\bibitem[Ehret et~al.(2022)Ehret, De~Truchis, Mazzolini, Morel, D’aspremont, Lauvaux, Duren, Cusworth, and Facciolo]{ehret2022global}
Thibaud Ehret, Aur{\'e}lien De~Truchis, Matthieu Mazzolini, Jean-Michel Morel, Alexandre D’aspremont, Thomas Lauvaux, Riley Duren, Daniel Cusworth, and Gabriele Facciolo.
\newblock Global tracking and quantification of oil and gas methane emissions from recurrent sentinel-2 imagery.
\newblock \emph{Environmental science \& technology}, 56\penalty0 (14):\penalty0 10517--10529, 2022.

\bibitem[Felzenszwalb et~al.(2008)Felzenszwalb, McAllester, and Ramanan]{4587597}
Pedro Felzenszwalb, David McAllester, and Deva Ramanan.
\newblock A discriminatively trained, multiscale, deformable part model.
\newblock In \emph{2008 IEEE Conference on Computer Vision and Pattern Recognition}, pages 1--8, 2008.

\bibitem[Felzenszwalb et~al.(2009)Felzenszwalb, Girshick, McAllester, and Ramanan]{felzenszwalb2009object}
Pedro~F. Felzenszwalb, Ross~B. Girshick, David McAllester, and Deva Ramanan.
\newblock Object detection with discriminatively trained part-based models.
\newblock \emph{IEEE transactions on pattern analysis and machine intelligence}, 32\penalty0 (9):\penalty0 1627--1645, 2009.

\bibitem[Freund and Schapire(1997)]{FREUND1997119}
Yoav Freund and Robert~E Schapire.
\newblock A decision-theoretic generalization of on-line learning and an application to boosting.
\newblock \emph{Journal of Computer and System Sciences}, 55\penalty0 (1):\penalty0 119--139, 1997.

\bibitem[Girshick et~al.(2014{\natexlab{a}})Girshick, Donahue, Darrell, and Malik]{rcnn}
Ross Girshick, Jeff Donahue, Trevor Darrell, and Jitendra Malik.
\newblock Rich feature hierarchies for accurate object detection and semantic segmentation.
\newblock In \emph{Proceedings of the IEEE conference on computer vision and pattern recognition}, pages 580--587, 2014{\natexlab{a}}.

\bibitem[Girshick et~al.(2014{\natexlab{b}})Girshick, Iandola, Darrell, and Malik]{DPM_CNN}
Ross~B. Girshick, Forrest~N. Iandola, Trevor Darrell, and Jitendra Malik.
\newblock Deformable part models are convolutional neural networks.
\newblock \emph{CoRR}, abs/1409.5403, 2014{\natexlab{b}}.

\bibitem[Han et~al.(2021)Han, Ding, Xue, and Xia]{Redet}
Jiaming Han, Jian Ding, Nan Xue, and Gui-Song Xia.
\newblock Redet: A rotation-equivariant detector for aerial object detection.
\newblock In \emph{Proceedings of the IEEE/CVF conference on computer vision and pattern recognition}, pages 2786--2795, 2021.

\bibitem[Haroon et~al.(2020)Haroon, Shahzad, and Fraz]{simd}
Muhammad Haroon, Muhammad Shahzad, and Muhammad~Moazam Fraz.
\newblock Multisized object detection using spaceborne optical imagery.
\newblock \emph{IEEE Journal of Selected Topics in Applied Earth Observations and Remote Sensing}, 13:\penalty0 3032--3046, 2020.

\bibitem[He et~al.(2017)He, Gkioxari, Doll{\'a}r, and Girshick]{maskrcnn}
Kaiming He, Georgia Gkioxari, Piotr Doll{\'a}r, and Ross Girshick.
\newblock Mask r-cnn.
\newblock In \emph{Proceedings of the IEEE international conference on computer vision}, pages 2961--2969, 2017.

\bibitem[Huang et~al.(2024)Huang, Feng, Liu, and Wang]{huang2024mutdetmutuallyoptimizingpretraining}
Ziyue Huang, Yongchao Feng, Qingjie Liu, and Yunhong Wang.
\newblock Mutdet: Mutually optimizing pre-training for remote sensing object detection, 2024.

\bibitem[{IEA}(2022)]{IEA2022}
{IEA}.
\newblock Methane and climate change – global methane tracker 2022 – analysis.
\newblock \url{https://www.iea.org/reports/global-methane-tracker-2022/methane-and-climate-change}, 2022.

\bibitem[{IGN - Institut national de l’information géographique et forestière}(2024)]{bdortho}
{IGN - Institut national de l’information géographique et forestière}.
\newblock {BD ORTHO®: L'image géographique du territoire national, la France vue du ciel}.
\newblock \url{https://geoservices.ign.fr/bdortho}, 2024.
\newblock Accessed: February 3, 2025.

\bibitem[{Info Spot 6-7}(2023)]{Spot6-7_2023}
{Info Spot 6-7}.
\newblock {Accès à l'Open Data Couvertures Spot 6-7 France}.
\newblock \url{https://www.geoportail.gouv.fr/actualites/observer-l-evolution-des-territoires-avec-spot}, 2023.
\newblock Accessed: June 10, 2025.

\bibitem[Jocher et~al.(2023)Jocher, Qiu, and Chaurasia]{yolov8}
Glenn Jocher, Jing Qiu, and Ayush Chaurasia.
\newblock {Ultralytics YOLO}, 2023.

\bibitem[Kang et~al.(2019)Kang, Liu, Wang, Yu, Feng, and Darrell]{fsrw}
Bingyi Kang, Zhuang Liu, Xin Wang, Fisher Yu, Jiashi Feng, and Trevor Darrell.
\newblock Few-shot object detection via feature reweighting.
\newblock In \emph{Proceedings of the IEEE/CVF International Conference on Computer Vision}, pages 8420--8429, 2019.

\bibitem[Khanam and Hussain(2024)]{yolov11}
Rahima Khanam and Muhammad Hussain.
\newblock Yolov11: An overview of the key architectural enhancements, 2024.

\bibitem[Law and Deng(2018)]{cornernet}
Hei Law and Jia Deng.
\newblock Cornernet: Detecting objects as paired keypoints.
\newblock In \emph{Proceedings of the European conference on computer vision}, pages 734--750, 2018.

\bibitem[Li et~al.(2017)Li, Cheng, Bu, and You]{obj_rs_2}
Ke Li, Gong Cheng, Shuhui Bu, and Xiong You.
\newblock Rotation-insensitive and context-augmented object detection in remote sensing images.
\newblock \emph{IEEE Transactions on Geoscience and Remote Sensing}, 56\penalty0 (4):\penalty0 2337--2348, 2017.

\bibitem[Li et~al.(2020)Li, Wan, Cheng, Meng, and Han]{LI2020296}
Ke Li, Gang Wan, Gong Cheng, Liqiu Meng, and Junwei Han.
\newblock Object detection in optical remote sensing images: A survey and a new benchmark.
\newblock \emph{ISPRS Journal of Photogrammetry and Remote Sensing}, 159:\penalty0 296--307, 2020.

\bibitem[Li et~al.(2023)Li, Hou, Zheng, Cheng, Yang, and Li]{li2023largeselectivekernelnetwork}
Yuxuan Li, Qibin Hou, Zhaohui Zheng, Ming-Ming Cheng, Jian Yang, and Xiang Li.
\newblock Large selective kernel network for remote sensing object detection, 2023.

\bibitem[Li et~al.(2024)Li, Li, Dai, Hou, Liu, Liu, Cheng, and Yang]{li2024lsknetfoundationlightweightbackbone}
Yuxuan Li, Xiang Li, Yimian Dai, Qibin Hou, Li Liu, Yongxiang Liu, Ming-Ming Cheng, and Jian Yang.
\newblock Lsknet: A foundation lightweight backbone for remote sensing, 2024.

\bibitem[Lin et~al.(2014)Lin, Maire, Belongie, Bourdev, Girshick, Hays, Perona, Ramanan, Doll{\'{a}}r, and Zitnick]{COCO}
Tsung{-}Yi Lin, Michael Maire, Serge~J. Belongie, Lubomir~D. Bourdev, Ross~B. Girshick, James Hays, Pietro Perona, Deva Ramanan, Piotr Doll{\'{a}}r, and C.~Lawrence Zitnick.
\newblock Microsoft {COCO:} common objects in context.
\newblock \emph{CoRR}, abs/1405.0312, 2014.

\bibitem[{Michael Fredenslund} et~al.(2023){Michael Fredenslund}, Gudmundsson, {Maria Falk}, and Scheutz]{danish_report}
Anders {Michael Fredenslund}, Einar Gudmundsson, Julie {Maria Falk}, and Charlotte Scheutz.
\newblock The danish national effort to minimise methane emissions from biogas plants.
\newblock \emph{Waste Management}, 157:\penalty0 321--329, 2023.

\bibitem[Mundhenk et~al.(2016)Mundhenk, Konjevod, Sakla, and Boakye]{COWC_CarsOverheadWithContext}
T.~Nathan Mundhenk, Goran Konjevod, Wesam~A. Sakla, and Kofi Boakye.
\newblock A large contextual dataset for classification, detection and counting of cars with deep learning.
\newblock In \emph{Computer Vision -- ECCV 2016}, pages 785--800, Cham, 2016. Springer International Publishing.

\bibitem[Murrugarra-LLerena et~al.(2025)Murrugarra-LLerena, Marques, and Jung]{murrugarrallerena2025gauchogaussiandistributionscholesky}
Jeffri Murrugarra-LLerena, Jose Henrique~Lima Marques, and Claudio~R. Jung.
\newblock Gaucho: Gaussian distributions with cholesky decomposition for oriented object detection, 2025.

\bibitem[on~Climate Change~(IPCC)(2023)]{Arias2021}
Intergovernmental~Panel on Climate Change~(IPCC).
\newblock \emph{Technical Summary}, page 35–144.
\newblock Cambridge University Press, 2023.

\bibitem[Ouerghi et~al.(2022)Ouerghi, Ehret, de~Franchis, Facciolo, Lauvaux, Meinhardt, and Morel]{ouerghi2022automatic}
Elyes Ouerghi, Thibaud Ehret, Carlo de Franchis, Gabriele Facciolo, Thomas Lauvaux, Enric Meinhardt, and Jean-Michel Morel.
\newblock Automatic methane plumes detection in time series of sentinel-5p l1b images.
\newblock \emph{ISPRS Annals of the Photogrammetry, Remote Sensing and Spatial Information Sciences}, 53:\penalty0 147--154, 2022.

\bibitem[Ouerghi et~al.(2023)Ouerghi, Ehret, Facciolo, Meinhardt, Morel, De~Franchis, and Lauvaux]{elyes}
Elyes Ouerghi, Thibaud Ehret, Gabriele Facciolo, Enric Meinhardt, Jean-Michel Morel, Carlo De~Franchis, and Thomas Lauvaux.
\newblock Methane plumes detection on prisma l1 images with the adjusted spectral matched filter and wind data.
\newblock In \emph{IGARSS 2023-2023 IEEE International Geoscience and Remote Sensing Symposium}, pages 7598--7601. IEEE, 2023.

\bibitem[Pandey et~al.(2019)Pandey, Gautam, Houweling, Van Der~Gon, Sadavarte, Borsdorff, Hasekamp, Landgraf, Tol, Van~Kempen, et~al.]{pandey2019satellite}
Sudhanshu Pandey, Ritesh Gautam, Sander Houweling, Hugo~Denier Van Der~Gon, Pankaj Sadavarte, Tobias Borsdorff, Otto Hasekamp, Jochen Landgraf, Paul Tol, Tim Van~Kempen, et~al.
\newblock Satellite observations reveal extreme methane leakage from a natural gas well blowout.
\newblock \emph{Proceedings of the National Academy of Sciences}, 116\penalty0 (52):\penalty0 26376--26381, 2019.

\bibitem[Qiao et~al.(2020)Qiao, Sun, and Zhang]{Pylon_Datatset}
Sijia Qiao, Yu Sun, and Haopeng Zhang.
\newblock Deep learning based electric pylon detection in remote sensing images.
\newblock \emph{Remote Sensing}, 12\penalty0 (11):\penalty0 1857, 2020.

\bibitem[Ramachandran et~al.(2024)Ramachandran, Irvin, Omara, Gautam, Meisenhelder, Rostami, Sheng, Ng, and Jackson]{GasInventory}
Neel Ramachandran, Jeremy Irvin, Mark Omara, Ritesh Gautam, Kelsey Meisenhelder, Erfan Rostami, Hao Sheng, Andrew~Y. Ng, and Robert~B. Jackson.
\newblock Deep learning for detecting and characterizing oil and gas well pads in satellite imagery.
\newblock \emph{Nature Communications}, 15\penalty0 (1):\penalty0 7036, 2024.

\bibitem[Razakarivony and Jurie(2016)]{VEDAI}
Sebastien Razakarivony and Frederic Jurie.
\newblock Vehicle detection in aerial imagery : A small target detection benchmark.
\newblock \emph{Journal of Visual Communication and Image Representation}, 34:\penalty0 187--203, 2016.

\bibitem[Redmon and Farhadi(2018)]{yolov3}
Joseph Redmon and Ali Farhadi.
\newblock Yolov3: An incremental improvement.
\newblock \emph{CoRR}, abs/1804.02767, 2018.

\bibitem[Redmon et~al.(2015)Redmon, Divvala, Girshick, and Farhadi]{yolov1}
Joseph Redmon, Santosh~Kumar Divvala, Ross~B. Girshick, and Ali Farhadi.
\newblock You only look once: Unified, real-time object detection.
\newblock \emph{CoRR}, abs/1506.02640, 2015.

\bibitem[Ren et~al.(2015)Ren, He, Girshick, and Sun]{fasterrcnn}
Shaoqing Ren, Kaiming He, Ross Girshick, and Jian Sun.
\newblock Faster r-cnn: Towards real-time object detection with region proposal networks.
\newblock \emph{Advances in neural information processing systems}, 28, 2015.

\bibitem[Ren et~al.(2016)Ren, He, Girshick, and Sun]{ren2016fasterrcnnrealtimeobject}
Shaoqing Ren, Kaiming He, Ross Girshick, and Jian Sun.
\newblock Faster r-cnn: Towards real-time object detection with region proposal networks, 2016.

\bibitem[Robinson et~al.(2021)Robinson, Chugg, Anderson, Ferres, and Ho]{PoultryInventory}
Caleb Robinson, Ben Chugg, Brandon~R. Anderson, Juan M.~Lavista Ferres, and Daniel~E. Ho.
\newblock Mapping industrial poultry operations at scale with deep learning and aerial imagery.
\newblock \emph{CoRR}, abs/2112.10988, 2021.

\bibitem[Robinson et~al.(2025)Robinson, Ortiz, Kim, Dodhia, Zolli, Nagaraju, Oakleaf, Kiesecker, and Lavista~Ferres]{microsoft}
Caleb Robinson, Anthony Ortiz, Allen Kim, Rahul Dodhia, Andrew Zolli, Shivaprakash~K Nagaraju, James Oakleaf, Joe Kiesecker, and Juan~M. Lavista~Ferres.
\newblock Global renewables watch: A temporal dataset of solar and wind energy derived from satellite imagery.
\newblock 2025.

\bibitem[Schapire(2003)]{schapire2003boosting}
Robert~E. Schapire.
\newblock The boosting approach to machine learning: An overview.
\newblock \emph{Nonlinear estimation and classification}, pages 149--171, 2003.

\bibitem[Shindell et~al.(2012)Shindell, Kuylenstierna, Vignati, van Dingenen, Amann, Klimont, Anenberg, Muller, Janssens-Maenhout, Raes, Schwartz, Faluvegi, Pozzoli, Kupiainen, Höglund-Isaksson, Emberson, Streets, Ramanathan, Hicks, Oanh, Milly, Williams, Demkine, and Fowler]{doi:10.1126/science.1210026}
Drew Shindell, Johan C.~I. Kuylenstierna, Elisabetta Vignati, Rita van Dingenen, Markus Amann, Zbigniew Klimont, Susan~C. Anenberg, Nicholas Muller, Greet Janssens-Maenhout, Frank Raes, Joel Schwartz, Greg Faluvegi, Luca Pozzoli, Kaarle Kupiainen, Lena Höglund-Isaksson, Lisa Emberson, David Streets, V. Ramanathan, Kevin Hicks, N.~T.~Kim Oanh, George Milly, Martin Williams, Volodymyr Demkine, and David Fowler.
\newblock Simultaneously mitigating near-term climate change and improving human health and food security.
\newblock \emph{Science}, 335\penalty0 (6065):\penalty0 183--189, 2012.

\bibitem[Stowell et~al.(2020)Stowell, Kelly, Tanner, Taylor, Jones, Geddes, and Chalstrey]{Stowell2020}
Dan Stowell, Jack Kelly, Damien Tanner, Jamie Taylor, Ethan Jones, James Geddes, and Ed Chalstrey.
\newblock A harmonised, high-coverage, open dataset of solar photovoltaic installations in the uk.
\newblock \emph{Scientific Data}, 7\penalty0 (1):\penalty0 394, 2020.

\bibitem[Sun et~al.(2021)Sun, Wang, Wang, Liu, and Fu]{SUN202150}
Xian Sun, Peijin Wang, Cheng Wang, Yingfei Liu, and Kun Fu.
\newblock Pbnet: Part-based convolutional neural network for complex composite object detection in remote sensing imagery.
\newblock \emph{ISPRS Journal of Photogrammetry and Remote Sensing}, 173:\penalty0 50--65, 2021.

\bibitem[Sun et~al.(2022)Sun, Wang, Yan, Xu, Wang, Diao, Chen, Li, Feng, Xu, et~al.]{fair1m}
Xian Sun, Peijin Wang, Zhiyuan Yan, Feng Xu, Ruiping Wang, Wenhui Diao, Jin Chen, Jihao Li, Yingchao Feng, Tao Xu, et~al.
\newblock Fair1m: A benchmark dataset for fine-grained object recognition in high-resolution remote sensing imagery.
\newblock \emph{ISPRS Journal of Photogrammetry and Remote Sensing}, 184:\penalty0 116--130, 2022.

\bibitem[Team(2020)]{SolarPanelInventory}
Scientific Data~Curation Team.
\newblock {A harmonised, high-coverage, open dataset of solar photovoltaic installations in the UK}.
\newblock \emph{Nature}, 2020.

\bibitem[Thorpe et~al.(2023)Thorpe, Green, Thompson, Brodrick, Chapman, Elder, Irakulis-Loitxate, Cusworth, Ayasse, Duren, et~al.]{thorpe2023attribution}
Andrew~K. Thorpe, Robert~O. Green, David~R. Thompson, Philip~G. Brodrick, John~W. Chapman, Clayton~D. Elder, Itziar Irakulis-Loitxate, Daniel~H. Cusworth, Alana~K. Ayasse, Riley~M. Duren, et~al.
\newblock Attribution of individual methane and carbon dioxide emission sources using emit observations from space.
\newblock \emph{Science advances}, 9\penalty0 (46):\penalty0 eadh2391, 2023.

\bibitem[Tian et~al.(2019)Tian, Shen, Chen, and He]{fcos}
Zhi Tian, Chunhua Shen, Hao Chen, and Tong He.
\newblock Fcos: Fully convolutional one-stage object detection.
\newblock In \emph{2019 IEEE/CVF International Conference on Computer Vision (ICCV)}, pages 9626--9635, 2019.

\bibitem[Wang et~al.(2022)Wang, Bochkovskiy, and Liao]{yolov7}
Chien-Yao Wang, Alexey Bochkovskiy, and Hong-Yuan~Mark Liao.
\newblock Yolov7: Trainable bag-of-freebies sets new state-of-the-art for real-time object detectors, 2022.

\bibitem[Wang et~al.(2024)Wang, Yeh, and Liao]{yolov9}
Chien-Yao Wang, I-Hau Yeh, and Hong-Yuan~Mark Liao.
\newblock Yolov9: Learning what you want to learn using programmable gradient information, 2024.

\bibitem[Williams et~al.(2025)Williams, Omara, Himmelberger, Zavala-Araiza, MacKay, Benmergui, Sargent, Wofsy, Hamburg, and Gautam]{acp-25-1513-2025}
James~P. Williams, Mark Omara, Anthony Himmelberger, Daniel Zavala-Araiza, Katlyn MacKay, Joshua Benmergui, Maryann Sargent, Steven~C. Wofsy, Steven~P. Hamburg, and Ritesh Gautam.
\newblock Small emission sources in aggregate disproportionately account for a large majority of total methane emissions from the us oil and gas sector.
\newblock \emph{Atmospheric Chemistry and Physics}, 25\penalty0 (3):\penalty0 1513--1532, 2025.

\bibitem[Wu et~al.(2019)Wu, Kirillov, Massa, Lo, and Girshick]{wu2019detectron2}
Yuxin Wu, Alexander Kirillov, Francisco Massa, Wan-Yen Lo, and Ross Girshick.
\newblock Detectron2.
\newblock \url{https://github.com/facebookresearch/detectron2}, 2019.

\bibitem[Xia et~al.(2018)Xia, Bai, Ding, Zhu, Belongie, Luo, Datcu, Pelillo, and Zhang]{dota}
Gui-Song Xia, Xiang Bai, Jian Ding, Zhen Zhu, Serge Belongie, Jiebo Luo, Mihai Datcu, Marcello Pelillo, and Liangpei Zhang.
\newblock Dota: A large-scale dataset for object detection in aerial images.
\newblock In \emph{Proceedings of the IEEE conference on computer vision and pattern recognition}, pages 3974--3983, 2018.

\bibitem[Xiao et~al.(2020)Xiao, Tian, Yu, Zhang, Liu, Du, and Lan]{obj_det_review_2}
Youzi Xiao, Zhiqiang Tian, Jiachen Yu, Yinshu Zhang, Shuai Liu, Shaoyi Du, and Xuguang Lan.
\newblock A review of object detection based on deep learning.
\newblock \emph{Multimedia Tools and Applications}, 79:\penalty0 23729--23791, 2020.

\bibitem[Xie et~al.(2021)Xie, Cheng, Wang, Yao, and Han]{Oriented_RCNN}
Xingxing Xie, Gong Cheng, Jiabao Wang, Xiwen Yao, and Junwei Han.
\newblock Oriented {R-CNN} for object detection.
\newblock \emph{CoRR}, abs/2108.05699, 2021.

\bibitem[Xu et~al.(2021)Xu, Fu, Wang, Wang, Chen, Xia, and Bai]{gliding_vertex}
Yongchao Xu, Mingtao Fu, Qimeng Wang, Yukang Wang, Kai Chen, Gui-Song Xia, and Xiang Bai.
\newblock Gliding vertex on the horizontal bounding box for multi-oriented object detection.
\newblock \emph{IEEE Transactions on Pattern Analysis and Machine Intelligence}, 43\penalty0 (4):\penalty0 1452--1459, 2021.

\bibitem[Yang and Yan(2020)]{csl}
Xue Yang and Junchi Yan.
\newblock Arbitrary-oriented object detection with circular smooth label.
\newblock In \emph{Computer Vision--ECCV 2020: 16th European Conference, Glasgow, UK, August 23--28, 2020, Proceedings, Part VIII 16}, pages 677--694. Springer, 2020.

\bibitem[Yang and Yan(2022)]{oriented_detection_1}
Xue Yang and Junchi Yan.
\newblock On the arbitrary-oriented object detection: Classification based approaches revisited.
\newblock \emph{International Journal of Computer Vision}, 130\penalty0 (5):\penalty0 1340--1365, 2022.

\bibitem[Yang et~al.(2021{\natexlab{a}})Yang, Yan, Feng, and He]{R3det}
Xue Yang, Junchi Yan, Ziming Feng, and Tao He.
\newblock R3det: Refined single-stage detector with feature refinement for rotating object.
\newblock In \emph{Proceedings of the AAAI conference on artificial intelligence}, pages 3163--3171, 2021{\natexlab{a}}.

\bibitem[Yang et~al.(2021{\natexlab{b}})Yang, Yan, Ming, Wang, Zhang, and Tian]{wasserstein_loss}
Xue Yang, Junchi Yan, Qi Ming, Wentao Wang, Xiaopeng Zhang, and Qi Tian.
\newblock Rethinking rotated object detection with gaussian wasserstein distance loss.
\newblock In \emph{International conference on machine learning}, pages 11830--11841. PMLR, 2021{\natexlab{b}}.

\bibitem[Yang et~al.(2021{\natexlab{c}})Yang, Yang, Yang, Ming, Wang, Tian, and Yan]{kld}
Xue Yang, Xiaojiang Yang, Jirui Yang, Qi Ming, Wentao Wang, Qi Tian, and Junchi Yan.
\newblock Learning high-precision bounding box for rotated object detection via kullback-leibler divergence.
\newblock \emph{Advances in Neural Information Processing Systems}, 34:\penalty0 18381--18394, 2021{\natexlab{c}}.

\bibitem[Yang et~al.(2019)Yang, Liu, Hu, Wang, and Lin]{reppoints}
Ze Yang, Shaohui Liu, Han Hu, Liwei Wang, and Stephen Lin.
\newblock Reppoints: Point set representation for object detection.
\newblock In \emph{2019 IEEE/CVF International Conference on Computer Vision (ICCV)}, pages 9656--9665, 2019.

\bibitem[Yu and Da(2023)]{psc}
Yi Yu and Feipeng Da.
\newblock Phase-shifting coder: Predicting accurate orientation in oriented object detection.
\newblock In \emph{Proceedings of the IEEE/CVF Conference on Computer Vision and Pattern Recognition}, pages 13354--13363, 2023.

\bibitem[Zhang et~al.(2014)Zhang, Donahue, Girshick, and Darrell]{ZhangDGD14}
Ning Zhang, Jeff Donahue, Ross~B. Girshick, and Trevor Darrell.
\newblock Part-based r-cnns for fine-grained category detection.
\newblock \emph{CoRR}, abs/1407.3867, 2014.

\bibitem[Zhao et~al.(2019)Zhao, Zheng, Xu, and Wu]{obj_det_review_1}
Zhong-Qiu Zhao, Peng Zheng, Shou-tao Xu, and Xindong Wu.
\newblock Object detection with deep learning: A review.
\newblock \emph{IEEE transactions on neural networks and learning systems}, 30\penalty0 (11):\penalty0 3212--3232, 2019.

\bibitem[Zhou et~al.(2022)Zhou, Yang, Zhang, Wang, Liu, Hou, Jiang, Liu, Yan, Lyu, Zhang, and Chen]{zhou2022mmrotate}
Yue Zhou, Xue Yang, Gefan Zhang, Jiabao Wang, Yanyi Liu, Liping Hou, Xue Jiang, Xingzhao Liu, Junchi Yan, Chengqi Lyu, Wenwei Zhang, and Kai Chen.
\newblock Mmrotate: A rotated object detection benchmark using pytorch.
\newblock In \emph{Proceedings of the 30th ACM International Conference on Multimedia}, 2022.

\bibitem[Ševo and Avramovic(2016)]{obj_rs_1}
Igor Ševo and Aleksej Avramovic.
\newblock Convolutional neural network based automatic object detection on aerial images.
\newblock \emph{IEEE Geoscience and Remote Sensing Letters}, 13:\penalty0 1--5, 2016.

\end{thebibliography}
